\documentclass[journal]{IEEEtran}
\ifCLASSINFOpdf
\else
\fi

\usepackage[font=footnotesize]{subfig}
\usepackage{algorithmic}
\usepackage[ruled,vlined]{algorithm2e}
\usepackage{lipsum}
\usepackage{comment}
\usepackage{supertabular}
\usepackage[normalem]{ulem}

\usepackage{float} \restylefloat{figure} %defines H option for floats

\usepackage[usenames]{color}

\usepackage{amsmath}
\def\Vop{\operatornamewithlimits{%
  \mathchoice{\vcenter{\hbox{\huge v}}}
             {\vcenter{\hbox{\Large v}}}
             {\mathrm{v}}
             {\mathrm{v}}}}
\usepackage{graphicx}
\usepackage{epstopdf}

\usepackage{graphicx}
\usepackage{rotating}
\usepackage{lscape}
\usepackage{bm} %defines \bm for bold greek letters, i.e. $\bm{\alpha}$
\usepackage{cite}
\usepackage{threeparttable}
\usepackage{tabularx}
\usepackage{booktabs}
\usepackage[table,xcdraw]{xcolor}
\usepackage{adjustbox}
\usepackage[utf8]{inputenc}

\usepackage{amsmath}   % From the American Mathematical Society
\usepackage{afterpage}
\usepackage[section]{placeins}
\usepackage{acronym}

\usepackage{graphicx}
\graphicspath{{./} {img/}}

\usepackage[hyphens]{url}
\usepackage{hyperref}
\usepackage{scalerel}
\usepackage{tikz}
\usetikzlibrary{svg.path}

\definecolor{orcidlogocol}{HTML}{A6CE39}
\tikzset{
  orcidlogo/.pic={
    \fill[orcidlogocol] svg{M256,128c0,70.7-57.3,128-128,128C57.3,256,0,198.7,0,128C0,57.3,57.3,0,128,0C198.7,0,256,57.3,256,128z};
    \fill[white] svg{M86.3,186.2H70.9V79.1h15.4v48.4V186.2z}
                 svg{M108.9,79.1h41.6c39.6,0,57,28.3,57,53.6c0,27.5-21.5,53.6-56.8,53.6h-41.8V79.1z M124.3,172.4h24.5c34.9,0,42.9-26.5,42.9-39.7c0-21.5-13.7-39.7-43.7-39.7h-23.7V172.4z}
                 svg{M88.7,56.8c0,5.5-4.5,10.1-10.1,10.1c-5.6,0-10.1-4.6-10.1-10.1c0-5.6,4.5-10.1,10.1-10.1C84.2,46.7,88.7,51.3,88.7,56.8z};
  }
}

\newcommand\orcidicon[1]{\href{https://orcid.org/#1}{\mbox{\scalerel*{
\begin{tikzpicture}[yscale=-1,transform shape]
\pic{orcidlogo};
\end{tikzpicture}
}{|}}}}

\begin{document}

\title{EBBINNOT: A Hardware-Efficient Hybrid Event-Frame Tracker for Stationary Dynamic Vision Sensors}

% \author{Deepak Singla, Vivek Mohan, Tarun Pulluri, Andres Ussa, Bharath Ramesh, Arindam Basu,~\IEEEmembership{Member,~IEEE}
% % , Jyotibdha Acharya,~\IEEEmembership{Member,~IEEE}, Tanay Karnik,~\IEEEmembership{Fellow,~IEEE}, Huichu Liu,~\IEEEmembership{Member,~IEEE}, Hai Li,~\IEEEmembership{Member,~IEEE}, Jae-sun Seo,~\IEEEmembership{Senior Member,~IEEE} and Chang Song
% % \thanks{Arindam Basu and Jyotibdha Acharya are with the School of Electrical and Electronic Engineering,
% % Nanyang Technological University, Singapore 639798 (e-mail:arindam.basu@ntu.edu.sg). Jae-sun Seo is with Arizona State University, Tempe, AZ 85287. Chang Song and Hai Li are with Duke University, Cary, NC 27519. Huichu Liu and Tanay Karnik are with Microarchitecture Research Lab, Intel corporation, Hillsboro, OR.}
% %
% % \thanks{Copyright (c) 2010 IEEE. Personal use of this material is
% % permitted. However, permission to use this material for any other
% % purposes must be obtained from the IEEE by sending an email to
% % pubs-permissions@ieee.org.} 
% }

\author{Vivek~Mohan*~\orcidicon{0000-0002-0248-6417},~\IEEEmembership{Member,~IEEE},
Deepak~Singla*~\orcidicon{0000-0001-7699-7079},  Tarun~Pulluri~\orcidicon{0000-0001-9798-3055}, Andres~Ussa~\orcidicon{0000-0001-8112-6681}, Pradeep~Kumar~Gopalakrishnan~\orcidicon{0000-0001-6597-6141},~\IEEEmembership{Senior Member,~IEEE}, Pao-Sheng~Sun, Bharath~Ramesh~\orcidicon{0000-0001-8230-3803},~\IEEEmembership{Member,~IEEE} and Arindam~Basu~\orcidicon{0000-0003-1035-8770},~\IEEEmembership{Senior Member,~IEEE}

%\thanks{**This work was supported through grant RG87/16 by MOE, Singapore}% <-this % stops a space

\thanks{V. Mohan, P. Gopalakrishnan are with the School of EEE, Nanyang Technological University, Singapore}

\thanks{D. Singla is currently with the Department of Bioengineering, University of California, Los Angeles}

\thanks{T. Pulluri, A. Ussa,  and B. Ramesh are with the N.1 Institute for Health, National University of Singapore, Singapore}

\thanks{P. S. Sun and A. Basu are with the Department of EE, City University of Hong Kong.}

\thanks{*\emph{V. Mohan and D. Singla are co-first authors}}
}


\maketitle

\begin{abstract}
As an alternative sensing paradigm, dynamic vision sensors (DVS) have been recently explored to tackle scenarios where conventional sensors result in high data rate and processing time. This paper presents a hybrid event-frame approach for detecting and tracking objects recorded by a stationary neuromorphic sensor, thereby exploiting the sparse DVS output in a low-power setting for traffic monitoring. Specifically, we propose a hardware efficient processing pipeline that optimizes memory and computational needs that enable long-term battery powered usage for IoT applications. To exploit the background removal property of a static DVS, we propose an event-based binary image creation that signals presence or absence of events in a frame duration. This reduces memory requirement and enables usage of simple algorithms like median filtering and connected component labeling for denoise and region proposal respectively. To overcome the fragmentation issue, a YOLO inspired neural network based detector and classifier to merge fragmented region proposals has been proposed. Finally, a new overlap based tracker was implemented, exploiting overlap between detections and tracks is proposed with heuristics to overcome occlusion. The proposed pipeline is evaluated with more than $5$ hours of traffic recording spanning three different locations on two different neuromorphic sensors (DVS and CeleX) and demonstrate similar performance. Compared to existing event-based feature trackers, our method provides similar accuracy while needing  $\approx{6}\times$ less computes. To the best of our knowledge, this is the first time a stationary DVS based traffic monitoring solution is extensively compared to simultaneously recorded RGB frame-based methods while showing tremendous promise by outperforming state-of-the-art deep learning solutions. The traffic dataset is publicly made available at: \textcolor{blue}{\emph{\url{https://nusneuromorphic.github.io/dataset/index.html}}}.
\end{abstract}

%Our proposed hybrid architecture had tracking accuracy (AUC = $0.45$) comparable to deep learning (DL) based trackers SiamMask (AUC = $0.477$) and SiamRPN++ (AUC = $0.473$) operating on simultaneously recorded RGB frames while requiring $\approx \textcolor{green}{12K}\times$ less computations. Compared to pure event based mean shift (AUC = $0.31$), our approach requires $\approx \textcolor{green}{12}\times$ more computations but provides much better accuracy. On the other hand, compared to event based feature tracking, our method provides similar accuracy while needing $\approx\textcolor{green}{5.89}\times$ less computes. Finally, we also evaluated our performance on two different NVS: DAVIS and CeleX and demonstrated similar gains.

\begin{flushleft}\label{AbbreviationsList}
\footnotesize{
\textbf{List of Abbreviations- } 
% EBBI: Event-based Binary Image, KF: Kalman Filter, OT: Overlap based Tracker, EBMS: Event-based Mean Shift, RP: Region Proposal, HIST RP: Histogram RP, CCL RP: Connected Component Labeling RP, 1B1C: 1-bit 1-channel image, 1B2C: 1-bit 2-channel image, BB: Bounding Box, NNDC RP: Neural Network Detector plus Classifier RP, GT: Ground Truth, IoU: Intersection-over-Union, AUC: Area under Curve, NMS: Non-maximal Suppression, EvFT RP: Event Feature Tracker \cite{Kostas-flow} based RP, EBBINN: EBBI + NNDC RP. 
% \textcolor{red}{
\begin{table}[H]
% \begin{tabular}{>{\color{red}}l >{\color{red}}l}
\begin{tabular}{ll}
% \begin{supertabular}{ll}
EBBI    & Event-based Binary Image                                           \\
KF      & Kalman Filter                                                      \\
OT      & Overlap-based Tracker                                              \\
EBMS    & Event-based Mean Shift                                             \\
RP      & Region Proposal                                                    \\
HIST RP & Histogram RP                                                       \\
CCL RP  & Connected Component Labeling RP                                    
\end{tabular}
\end{table}
% }
% \textcolor{red}{
\begin{table}[H]
% \begin{tabular}{>{\color{red}}l >{\color{red}}l}
\begin{tabular}{ll}
\\
1B1C    & 1-bit 1-channel image                                              \\
1B2C    & 1-bit 2-channels image                                             \\
BB      & Bounding Box                                                       \\
NNDC RP & Neural Network Detector plus Classifier RP                         \\
GT      & Ground Truth                                                       \\
IoU     & Intersection-over-Union                                            \\
AUC     & Area Under Curve                                                   \\
NMS     & Non-Maximal Suppression                                            \\
EvFT    & Event Feature Tracker based RP \\
EBBINN  & EBBI + NNDC RP                                                    
% \end{supertabular}
\end{tabular}
\end{table}
}%FOV: Field-of-view
% }
\end{flushleft}

% \makeatletter{\renewcommand*{\@makefnmark}{}
% \footnotetext{Manuscript received on Nov 14 and revised on Nov 21, 2017.}\makeatother}



% \fi
\IEEEpeerreviewmaketitle

\section{Introduction}
Conventional vision sensors are ubiquitously used around the world for internet of things (IoT) applications. However, many IoT practical applications such as traffic monitoring do not require very high precision tracking--rather, it is more important to reduce false positives with minimal processing and latency. Operating on a retina-inspired principle, dynamic vision sensors (DVS) \cite{Lichtsteiner2008} provide advantages of ideal sampling due to sensing based on change detection. They also provide low data rates, high dynamic range and high effective frame rate~\cite{nvs-review,basu2018low,DVS-review}. It has largely been touted to be useful for object tracking by various works~\cite{ni2012, liu2016combined, camunas2017event, delbruck2013robotic, ramesh2018long}. However, event driven tracking requires very stringent denoise operations to reduce false positives--often found to be quite difficult to achieve. 
\par
% Does our predator-prey papers do this comparison? I thought we did such comparison, Papers: cite as \cite{moeys2016steering, moeys2018pred18} D. P. Moeys, F. Corradi, E. Kerr, P. Vance, G. Das, D. Neil, D. Kerr, and T. Delbrück, "Steering a predator robot using a mixed frame/event-driven convolutional neural network," in 2016 Second International Conference on Event-based Control, Communication, and Signal Processing (EBCCSP), 2016, pp. 1–8. D. P. Moeys, D. Neil, F. Corradi, E. Kerr, P. Vance, G. Das, S. A. Coleman, T. M. McGinnity, D. Kerr, and T. Delbruck, "PRED18: Dataset and Further Experiments with DAVIS Event Camera in Predator-Prey Robot Chasing," in IEEE Fourth International Conference on Event-Based Control, Communication and Signal Processing (EBCCSP 2018), Perpignan, France, 2018. 
While DVS reduces the data rate compared to a conventional RGB sensor, it is also necessary to develop a full processing pipeline of low complexity operators that can result in energy efficient hardware for deployment. Moreover, with the massive growth in Deep learning (DL) solutions, it is essential to ask the question of how well does a DVS perform in object detection and tracking as compared to regular camera output processed by DL frameworks. Comprehensive comparisons with a regular RGB image sensor on the same application, particularly for stationary DVS, is largely unavailable. Earlier comparisons were either for moving DVS or for DVS +  low-resolution, grayscale images as shown in \cite{moeys2016steering, moeys2018pred18} while arguably high resolution RGB images might have more information for deep learning (DL) based methods. This work addresses this gap by using simultaneously recorded RGB and event data for a traffic monitoring application. 
\par
In this work, we show that in applications such as traffic monitoring with stationary DVS, the change detection property of DVS can enable high accuracy detection and tracking when combined with simple DL techniques of significantly lower complexity than conventional ones\cite{redmon2018yolov3}. In particular, we propose a new processing pipeline for stationary neuromorphic cameras that uses a hybrid approach of creating event-based binary image (EBBI) involving time collapsing and intensity quantization of event stream. This also enables duty cycled operation of the DVS making it compatible with commercial off-the shelf hardware such as microcontroller units and FPGA for IoT that rely on duty cycling for reducing energy. The use of simple image processing-based filtering techniques for denoising the EBBI, with noise suppression comparable to conventional event-based noise filtering approaches such as nearest neighbour filter~\cite{delbruck2008frame,fnins2018}, sometimes also referred to as background activity filter. These denoised EBBI frames require lower memory, making them suitable for implementation while simplifying the detection and tracking components in the proposed pipeline.

\par

The rest of the paper is organized as follows. Related works are discussed in the next section. Section \ref{sec:materials} describes the proposed EBBINNOT framework. Section \ref{sec:mod_results} presents the performance and computation complexity of each block as well as the whole pipeline and compares them with relevant baselines such as histogram based region proposal (HIST RP), Kalman filter (KF) based tracking, pure event-based mean shift (EBMS) tracking \cite{delbruck2013robotic}, and pure RGB frames followed by DL-based tracking. This is followed by a section that discusses the main results and also shows that our approach is DVS independent and yields expected results with two commercially available DAVIS240C\cite{brandli2014240} and CeleX~\cite{guo2017live}. Finally, we conclude in the last section.
% We compare our proposed EBBI + NN-DC + OT approach with the following comparable baselines:
% \begin{itemize}
% \item (EBBI + NN-DC + OT) Vs. (EBBI + Hist RP + OT)
% \item (EBBI + NN-DC + OT) Vs. (EBBI + NN-DC + KF)
% \item (EBBI + NN-DC + OT) Vs. Conventional image sensor with DL based tracker.
% \end{itemize}

\section{Related work}
\label{sec:related}

Real-time object tracking consists of initializing candidate regions for objects in the scene, assigning these unique identifiers and following their transition. It is a common requirement to further perform classification over the tracked object. These capabilities of object tracking and classification are valuable in applications like human-computer interaction \cite{Jacob2003}, traffic control \cite{Hsieh2006}, autonomous vehicles \cite{Muresan2019}, medical imaging \cite{Meijering2006} or video security and surveillance \cite{Hampapur2005}. Current methodologies for surveillance tasks use standard cameras that acquire images or frames at a fixed rate regardless of scene dynamics. Consequently, background subtraction used to retrieve candidate regions-of-interest for tracking is a computationally intensive step, which is also affected by changes in lighting \cite{Piccardi2004}. On the other hand, deployment of cameras with higher frame rate involves a drastic increase in power requirements \cite{Barnich2011}, besides increased demands in memory and bandwidth transmission. Therefore, frame-based paradigm tends to be intractable for embedded platforms/remote surveillance applications \cite{Basu2018, Cohen2018, frame1, frame2, frame3}. 
\par

As an emerging alternative to standard cameras, event cameras acquire information of a scene in an asynchronous and pixel independent manner, where each of them react and transmit data only when intensity variation is observed. This provides a steady stream of events with a very high temporal resolution (microsecond) at low-power ($5-14mW$), reducing redundancy in the data with improved dynamic range due to the local processing paradigm. An event-by-event approach is dominantly seen in the literature for object tracking and detection using neuromorphic vision sensors \cite{Lagorce2014, Glover2017, Ramesh2019, event1, event2}. The aim of these methods is to create an object representation based on a set of incoming events and updating it dynamically when events are triggered. Although these methods can be effective for specific applications, they often require high parametrization \cite{Lagorce2014, Glover2017} or are not effective for tracking multiple objects \cite{Ramesh2019}. Similar to the above works, \cite{Dardelet2018} is an event-by-event approach for object tracking applications that performs a continuous event-based estimation of velocity using a Bayesian descriptor. Another example is \cite{ramesh2018long}, which proposes event-based tracking and detection for general scenes using a discriminative classification system and a sliding window approach. While these methods work intuitively for objects with different shapes and sizes, and even obtain good tracking results, they have not been implemented under a real-time operation requirement. Current event-based processing algorithms also require a significant amount of memory and processing due to noise related events.

\par

% In particular, there is no significant need for background modeling, since a static event camera will only generate events corresponding to moving objects, thereby naturally facilitating tracker initialization. All these features are well suited for visual tracking applications but demand the use of algorithms designed to handle asynchronous events.
\par

In contrast to the above methods, an aggregation of incoming events can be considered at fixed intervals instead of processing events as they arrive. This produces a more obvious representation of the scene (a ``frame''), and allows an easier coupling with traditional feature extraction and classification approaches \cite{Ni2015, Hinz2017, Iacono2018}. In \cite{Hinz2017}, asynchronous event data is captured at different time intervals, such as $10$ ms and $20$ ms, to obtain relevant motion and salient information. Then, clustering algorithms and Kalman filter are applied for detection and tracking, achieving good performance under limited settings. Other examples of event-based frames along with variations in sampling frequency and recognition techniques are \cite{Ni2015, Iacono2018}, which show the potential of this approach for detection. Taking an important step forward for real-time and embedded applications, we leverage the low-latency and high dynamic range of event cameras for static surveillance applications. Some recent work has demonstrated custom in-memory computing chips for simple region proposals\cite{Sumon2021} (without neural networks) on EBBI with high energy efficiency but low accuracies. On the other hand, there are many published work using embedded systems such as FPGA \cite{Gschwend2020,Jahanshahi2021}, MPSOC\cite{Minakova2020} or custom neuromorphic chips\cite{Lin2018} for implementing deep neural networks but not for visual tracking (though see \cite{Ullah2020} for one example). Given the huge variety of embedded systems available, in this work we do not report specific energy or latency figures corresponding to a specific implementation but rather compare the computational and memory complexity of different current approaches with our proposed low-complexity algorithm. The works in \cite{Litzenberger2006_DSP,Gritsch2008,Gritsch2009} are most closely related with our work in that they also use an embedded system based on DVS for traffic monitoring. We have compared our work with an updated version of their mean shift tracking algorithm, EBMS, in Sec. \ref{sec:mod_results}. Also, compared to these works which classified objects in only two categories, we have a finer classification with four categories. Finally, we also critically analyze tracking performance using F-1 measures and expected average overlap (EAO), and compare our work with DL based trackers operating on simultaneously recorded RGB images. Some other recent work on using DVS for tracking use attractor neural networks\cite{Leideng-tracking} or combine frames and events\cite{dashnet-arxiv}. However, they both require a separate object detection module and are only meant for single object tracking in contrast to our work which is more general with multi-object detection and tracking.

\par
\begin{figure} []
\centering     %%% not \center
\includegraphics[width=0.45\textwidth]{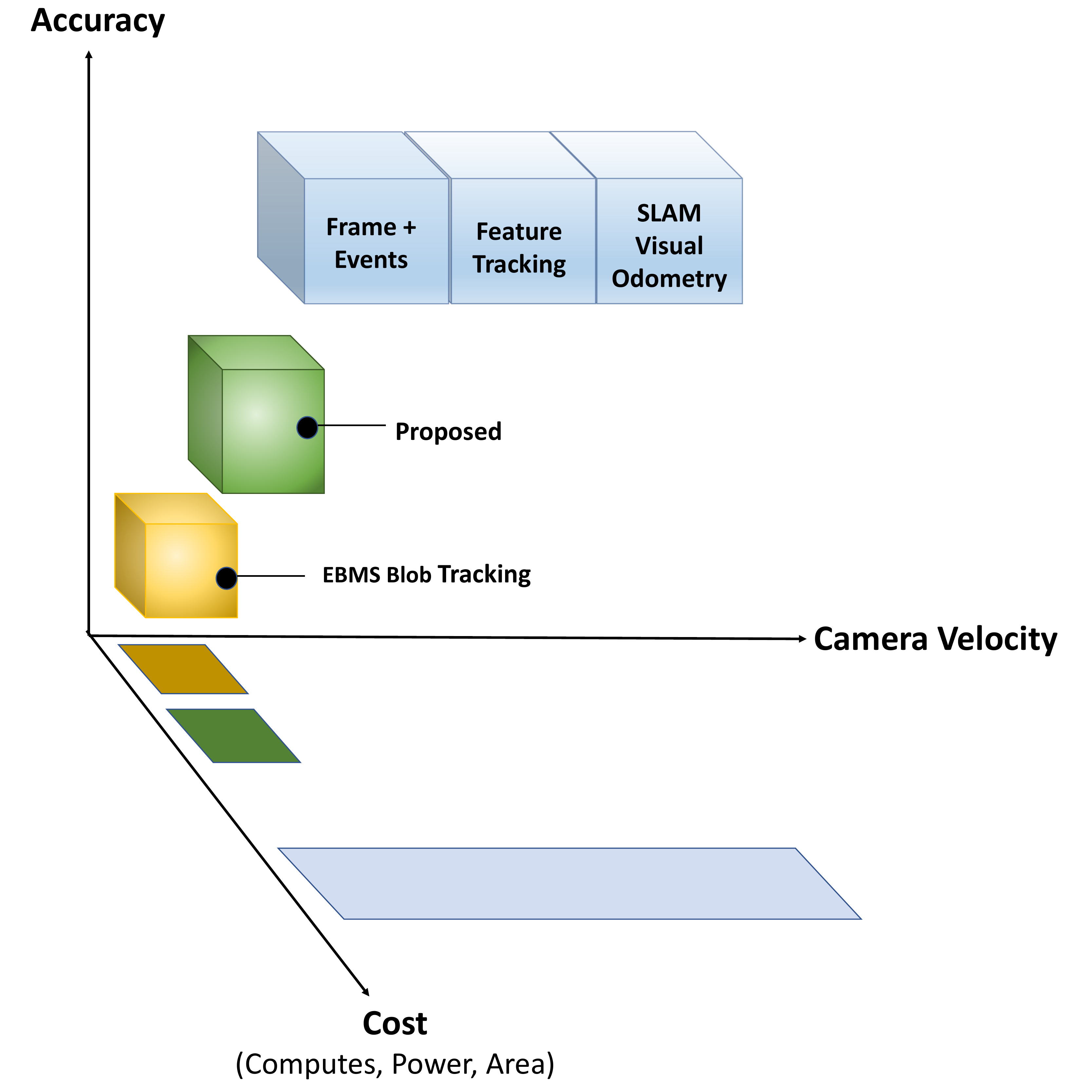}
 \caption{
%  \textcolor{red}
 { A generic overview of our proposed work in comparison to other trackers for DVS data. The proposed work is less computationally complex than event-based algorithms for optical flow based feature tracker as in \cite{Kostas-flow} or visual odometry\cite{Davide-slam}, while tracking with accuracy much higher than cluster trackers\cite{delbruck2013robotic} for stationary DVS scenario.}}
\label{fig:motivation}
\end{figure}
Figure \ref{fig:motivation} puts our work in the context of other reported works for tracking using DVS data. Some event based methods proposed for tracking output of DVS, such as cluster trackers\cite{delbruck2013robotic} are very computationally cheap but the accuracy is not very high even for stationary DVS scenario. On the other hand, more sophisticated methods such as those used for optical flow\cite{Kostas-flow} or SLAM\cite{Davide-slam} are very accurate in tracking even for moving DVS but with orders of magnitude increase in computations. The method proposed in this paper strikes a compromise with computational complexity close to cluster trackers but accuracy close to the best event based trackers for stationary DVS.

\par
Since our proposed solution combines EBBI, neural network (NN) based region proposal and an overlap tracker (OT), we refer to it as EBBINNOT.  An earlier version of this work was presented in \cite{AuthorsEBBIOT}--however, the histogram region proposal used in \cite{AuthorsEBBIOT} suffered from inaccurately sized and fragmented regions. The contributions of this paper are: 
\begin{itemize}
    \item A hybrid neural network based detector-classifier (NNDC) flow for merging fragmented object bounding boxes and object aware false positive suppression caused by EBBI frame generation. The NNDC rectified bounding boxes can then be fed to a tracker.
    \item A computationally inexpensive tracker that exploits overlap between predicted bounding boxes of the tracker owing to the fast sampling enabled by DVS.
    \item Providing a new stationary DVS dataset of more than $5$ hours for tracking vehicles and pedestrians in city traffic conditions.
\end{itemize}

\begin{figure}[t]
\centering     %%% not \center
\includegraphics[width=\linewidth]{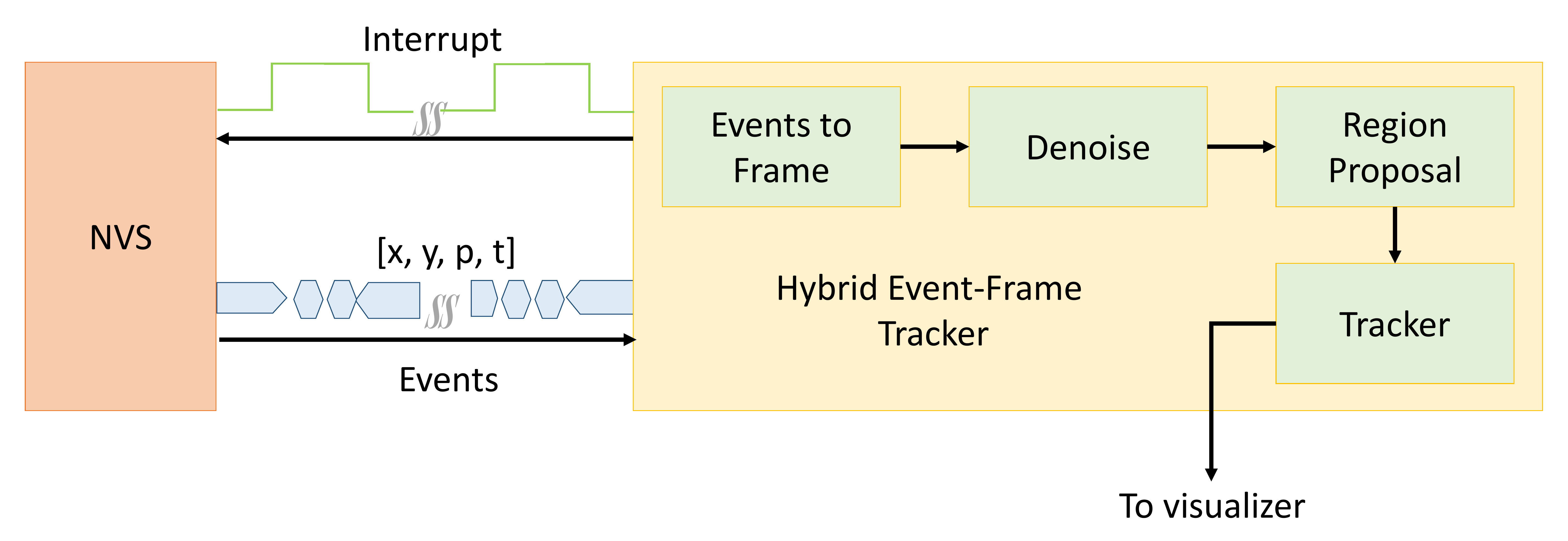}
 \caption{Generalized block diagram for EBBINNOT}
\label{fig:gen_blockdiagram}
\end{figure}

\section{EBBINNOT: Algorithmic Framework}
\label{sec:materials}

Figure~\ref{fig:gen_blockdiagram} shows a block diagram of EBBINNOT depicting the major blocks in the processing pipeline as well as the possibility of duty cycled interface with a DVS. It is to be noted that such hybrid approaches are becoming popular recently~\cite{sironi2018hats} and supporting hardware solutions are also being released~\cite{pei2019towards}. 

\begin{figure*}
\centering
\resizebox{\textwidth}{!}{
\includegraphics[]{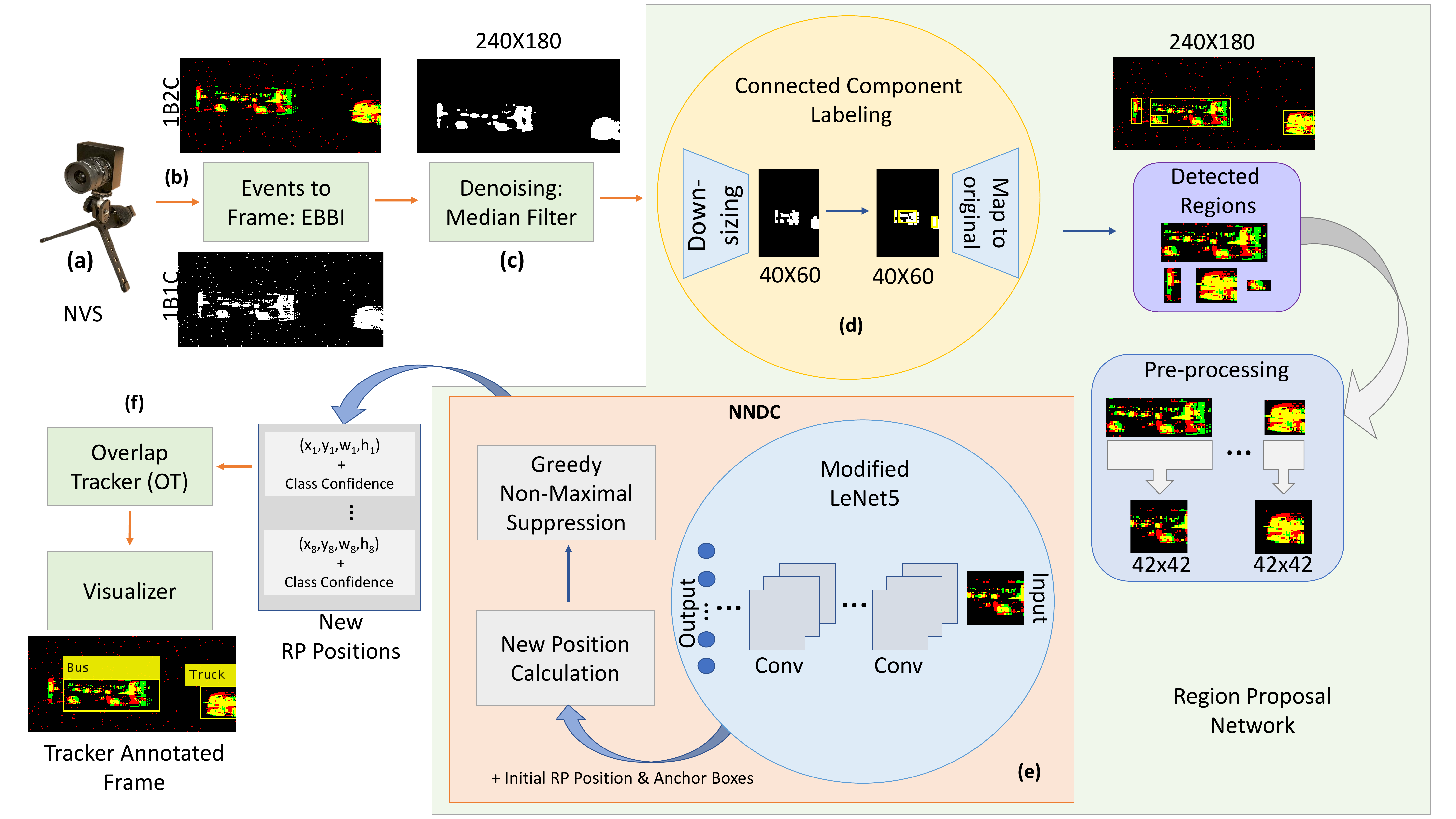}
}
\caption{Detailed block diagram of the EBBINNOT pipeline for $240\times 180$ sensor like DAVIS. For larger sensor like CeleX, the image was downsampled appropriately by dropping lower address bits of the events, thus mapping multiple sensor pixels to same image location. Input events from the (a) DVS is converted to a binary image in (b) EBBI module (ON and OFF events in 1B2C is shown by two different colours) followed by (c) median filtering for noise removal. The region proposal (RP) consists of (d) connected component labelling and (e) NNDC blocks. The last block is the (f) tracker}
\label{fig:pipeline}
\end{figure*}

The DVS camera, unlike traditional image sensors with fixed frame rate, operates by detecting temporal contrast (or change in log-intensity) at all pixels in parallel. If the change is larger than a threshold, it generates an asynchronous digital pulse or spike or event with a timestamp and a pixel location associated to it. Further, a polarity is assigned to each and every event according to the direction (increase or decrease) of contrast variation. This type of signalling is referred to as address event representation (AER). These changes in the format of data produced hence require a paradigm shift in the algorithms required for input processing in various applications, opening up a new avenue in engineering\cite{DVS-review}.
\par
Mathematically, an event can be modeled as $[e_i = (x_i, y_i, t_i, p_i)]$ where $(x_i,y_i)$ represents the event location or address on the sensor array, $t_i$ represents the timestamp of the event and $p_i$ represents the polarity associated to it~\cite{brandli2014live}. The associated timestamps to each event have microsecond resolution with quick readout rates ranging from $2$ MHz to $1200$ MHz. The event camera has an in-built invariance to illumination, since it detects temporal contrast change  largely cancelling out the effect of scene illumination. In short, the variation in log intensity represents the variation in reflectance due to the movement of the objects in the view. 
\par
The proposed EBBINNOT system consists of three major blocks (Fig. \ref{fig:pipeline}): EBBI and noise filtering, region proposal network (RPN) and tracking described in details below. A full list of parameters used in this paper can be found in Sec. 1 of the \emph{Supplementary Material}. \cite{ebbinnot-supplement}.
\subsection{Event Data Pre-processing}
\subsubsection{Event-Based Frame Generation}
\label{subsec:ebframegen}
In this work, we propose to aggregate events occurring within a specified time-interval (denoted by $t_F$ for frame time) into two types of temporally collapsed images. First, a \textit{single channel binary image} (1-bit, 1-channel image denoted as 1B1C in Fig. \ref{fig:pipeline}) was created by considering a pixel to be activated for any event mapping to the pixel location occurring within the interval $t_F$, irrespective of the polarity of the event and the event count for that pixel location. Mathematically, the equation for the (i,j)-th pixel in the k-th frame $I^k$ can be written as shown in eq. \ref{eq:ebbi}.
\begin{equation}
\label{eq:ebbi}
I^k(i,j)=
\begin{cases}
1, &\text{ if }\exists (i,j,t,p),p\epsilon\{0,1\},(k-1)t_F\leq t< kt_F\\
0, &\text{ otherwise}
\end{cases}
\end{equation}
\par
Second, a \textit{dual channel binary image} (1-bit, 2-channel image denoted as 1B2C in Fig. \ref{fig:pipeline}) was obtained in the same way as in case of 1B1C, with the exception that events corresponding to two polarities are written separately, with one channel consisting of ON events and the other consisting of OFF events. Note that 1B1C can be obtained by logical OR of the two 1B2C images--however, in practice, it is better to create the two images simultaneously to avoid further delays due to memory access. These methods also minimize high frequency noise characterized by abnormal firing rates in some of the pixel similar to refractory filtering for DVS \cite{garrick_noise,fnins2018} with refractory time $t_F$.
\par
Note that this is different from the downsampling methods in \cite{downsample-andre} where the total number of events in frame duration is counted to create a multi-bit image which has been shown to be not as informative as 1B2C for classification\cite{singla2020hynna}. Event-count based images may be thresholded to arrive at these EBBIs--counterintuitively, these thresholded representations produce superior results. Moreover, they have the advantage of being hardware friendly and amenable to processing via application of simple morphological operators\cite{gonzalez2008}. Akin to how visual information exits the occipital lobe into two distinct visual systems composed of \textit{what and where pathways}~\cite{merolla2014million}, in our work, the region proposal network for locating the object is comparable to the \textit{where pathway}. 
\par
Finally, while the hardware implementation is a future work for us, this proposed method of EBBI allows the processor (Fig. \ref{fig:gen_blockdiagram}) to be duty cycled since the DVS can act as a memory and retain the addresses of all events triggered in the interval till the processor wakes up, reads and resets it. Such fixed time duration based wake-up is the commonly available modality in all embedded hardware for IoT applications. Other methods of frame generation such as \cite{roshambo,jyoti-frontiers} relying on fixed event count, are unsuitable when there are multiple objects in the frame with varying sizes$/$speeds and would also need special hardware.  Some DVS imagers do have a provision to generate frames\cite{Brandli2014}; however, such multi-bit frames would require more complex interfacing circuits and larger memory. Moreover, our proposed EBBI is more suited to in-memory computing (IMC) based very low-energy image processing chips that are our goal. \cite{Sumon2021} presents an example of one of our first efforts in this direction that exhibits extremely high energy efficiency. Future work in this direction will explore adaptive time frames based on area-event number approaches\cite{Liu2018}. Lastly, we use $t_F=66$ ms in this work, but have seen the general concept works for a range of $t_F$ varying from $30-120$ ms. Even lower values of $t_F$ might be needed for tracking faster objects at the expense of power dissipation, while going to $t_F>120$ ms led to very high motion blur in our application  \cite{liu2019event}.
\subsubsection{Noise Filtering}
\label{subsec:noisefilt}
A conventional event-based filtering for an event stream from the DVS involves a combination of refractory filtering and an event-based background activity filter (BAF)~\cite{BAF_tobi} or nearest neighbour filter (NN-Filter) which passes events occurring within a correlation time in the neighbourhood of the event. %For events in an $A\times B$ sensor dimension, represented by $B_t$ bits per timestamp, a $p\times p$ NN-Filter, performs $p^2 - 1$ counter increments and comparisons besides a memory write for $B_t$ bits. The total computes and memory required when NN-Filtering is performed for an average of $\overline{n}$ events per frame was obtained as follows in~\cite{AuthorsEBBIOT}:
%\begin{align}
%\label{eq:NN_Filt}
%C_{NN-Filter} =& (2 (p^2 - 1) + B_t) \times \overline{n} \notag\\
%M_{NN-Filter} =& B_{t} \times A \times B
%\end{align}
%Note that $\overline{n}=\beta \times \alpha \times A\times B$ where $\beta(>1)$ denotes the average number of times a pixel fires in duration $t_F$ and $\alpha$ is the number of active pixels in the frame.
The creation of EBBI already provides refractory filtering as discussed earlier. It also enables us to leverage the use of median filter, a standard image processing tool, which ensures noise removal by replacing a pixel with the median value in its $p \times p$ neighborhood, while preserving the object edges~ \cite{gonzalez2008}. Formally, the (i,j)-th pixel in the filtered image $I_f$ can be denoted as shown below in eq. \ref{eq:medianfilt}:
\begin{equation}
\label{eq:medianfilt}
I_f(i,j)=
\begin{cases}
1, &\text{ if } \sum\limits_{n=-\frac{p-1}{2}}^{\frac{p-1}{2}}\sum\limits_{m=-\frac{p-1}{2}}^{\frac{p-1}{2}}I(i-m,j-n)>\lfloor\frac{p^2}{2}\rfloor\\
0, &\text{ otherwise}
\end{cases}
\end{equation}
where the superscript for frame index is dropped for convenience. Mathematically, this is related to the conventional event-based noise filters in the sense that the threshold for passing an event is $\lfloor p^2/2\rfloor$ for median filter while the same is $1$ for BAF or NN-filt. For example, a $p=3$ as used in this work, median filter would require the support of at least 4 events in the 3x3 neighborhood where as BAF requires support of only $1$ event within the correlation time in the same neighborhood. Hence, our proposed method is a more stringent noise filter.

% For removal of spurious noise due to pixel firing which roughly translates to salt and pepper type of noise in a 1B1C EBBI, a median filter performs an equivalent of $p^2$ counter increments for every $1$ and  $\lfloor p^2/2\rfloor$ comparisons, besides memory writes for creating the filtered EBBI. The average number of computations per $A \times B$ frame using a median filter on an EBBI is given by:
% \begin{align}
% \label{eq:MedianFilt}
% C_{Median-Filter} =& (\alpha p^2 + 2)\times A \times B \notag\\
% M_{Median-Filter} =& 2 \times A \times B
% \end{align}
% where the memory requirement is to store the raw and filtered 1B1C EBBI frames.

Assuming scenes with $10\%$ pixel activation recorded with a DAVIS 240 sensor\cite{brandli2014240} (sensor dimension of $240\times180$), calculations presented in \cite{AuthorsEBBIOT} conservatively estimates the number of computes for NN-Filter and Median Filter as $C_{\rm{NN-Filter}}\approx 276.4$ Kops/frame and $C_{\rm{Median-Filter}}=125.2$ Kops/frame (for window of $3\times3$) respectively. It was reported that the median filter approach requires nearly $8\times$ lesser memory than a conventional NN-filter. In Section \ref{sec:mod_results} we further show that the performance of the EBBI with median filtering is at par with the much more expensive NN-Filter.
\begin{figure*}[t]
\centering
% \resizebox{\textwidth}{!}{
\includegraphics[width=0.8\textwidth]{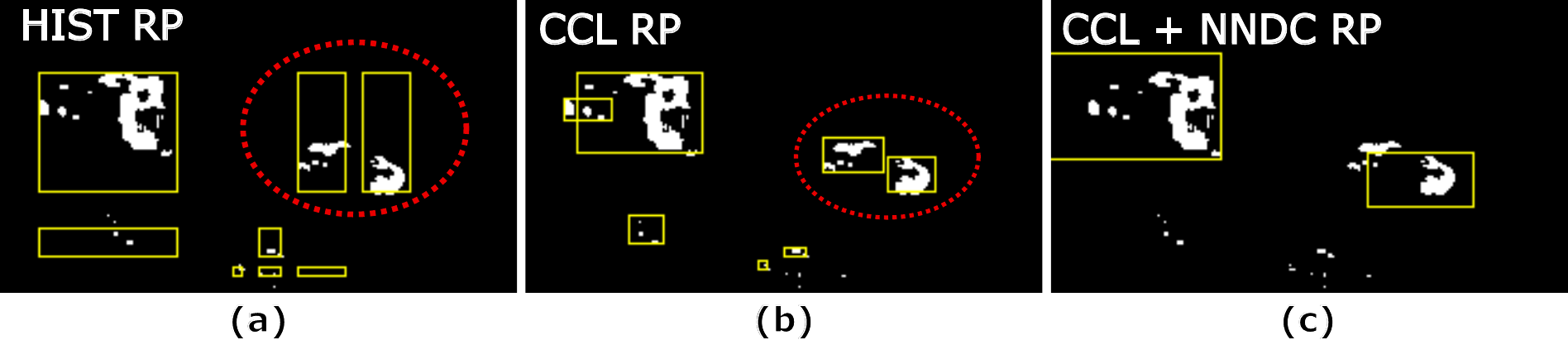}
% }
\caption {Comparison of different region proposal methods: (a) HIST RP presents the problem of enlarged and fragmented bounding box due to the presence of bigger object, (b) CCL RP resolves the inaccurate bounding box issue posed by (a) however object bounding box fragmentation is still observed, (c) CCL+NNDC RP resolves fragmentation problem and removes unwanted bounding boxes}
%\caption{Comparison of different RPs: (a) HIST RP: Enlarged and fragmented bounding box due to the presence of bigger object. (b) CCL RP: Solved big bounding box problem but object fragmentation observed. (c) CCL + NNDC RP: Solved fragmentation problem and removed unwanted boxes }
\label{fig:rps_comparison}
\end{figure*}

\subsection{Region Proposal Networks}
A crucial step to understand the visual scene involves the detection of salient visual cues. The role of SC behind the natural vision~\cite{veale2017visual,white2017superior} is a perfect example for detection. Natural vision pathway astounds researchers mostly because of its speed and efficiency, and SC proves efficient here by obtaining salient objects from a low spatial resolution version of the input~\cite{yohanandan2018saliency}. Surprisingly, the low resolution achromatic images allows better performance and faster response due to less computes.  Inspired by these, we propose to use a low-resolution version of 1B1C images in this work for the first phase of region proposal as described next. 

\subsubsection{Move from HIST RP to CCL RP}
%The projection of event information into EBBI and rejection of background by a stationary DVS provides us an opportunity to use well-known frame-based simple operators like edge detection and thresholding. To understand the context in the frame, learning the distribution of active pixels in an already foreground background separated EBBI is the key.
Histogram based RP (HIST RP) explored in~\cite{AuthorsEBBIOT, Authorsbmvc19,evtsurface-frontiers}, extracts one-dimensional (1-D) X and Y histograms by summing up all the active pixels along the respective axis. These histogram distributions can then be easily analyzed and the consecutive entries higher than some threshold can be used to locate the probable object locations back in 2-D. 

%However, operating this algorithm on the image at original sensor resolution can likely yield two or more areas for the fragmented images (e.g. glass windows in cars do not generate events and lead to fragmented clusters of events representing a car as shown in Fig. \ref{fig:rps_comparison}(a)). An appropriately chosen down-scaled version of the same image merges most of the objects. Further, a second run to weed out the false regions is done by checking the presence of active pixels in the proposed regions from the previous step. 
However, this algorithm suffers from the shortcoming of projecting back from 1-D to 2-D where the box for the smaller object gets affected in the presence of a bigger object (shown in Fig.~\ref{fig:rps_comparison}). A tight bounding box (BB) is required for a better understanding of the object in the classification stage. 

Therefore, instead of using 1-D projections, a natural choice is to use the morphological 2-D operator like connected component labeling (CCL RP). CCL RP relies on the connectivity of a target pixel with its surrounding eight pixels, called 8-connectivity neighbours. A two-pass algorithm of CCL reviewed in~\cite{he2017connected,walczyk2010comparative} and proposed for operation on 1B1C EBBI in~\cite{singla2020hynna}, produces tight BBs for an effective classification process. This algorithm relies on the equivalent label in the 8-connectivity neighborhood and continuously updates the BB corners of each and every pixel using the equivalent label during its two raster scans. Applied on a downsized version of EBBI for the same reason as HIST RP, this RP also keeps the computes in control. The downsizing is also a great example of exploration of low spatial resolution saliency detection aspect of the human visual system. The downsizing is done by scaling factors $s_1$ and $s_2$ as shown below in eq. \ref{eq:image_scale}:
\begin{align}
\label{eq:image_scale}
    I^{s_1,s_2}(i,j)=\Vop_{m=0,n=0}^{m=s_1-1,n=s_2-1}I(is_1+m,js_2+n)\notag\\
    i<\left\lfloor A/s_1\right\rfloor, j<\left\lfloor B/s_2\right\rfloor
\end{align}
where $I(i,j)\epsilon \{0,1\}$, $s_1$, $s_2$ are rescaling factors along X and Y axis, $A\times B$ is the sensor dimension and $\Vop$ represents the logical-OR operation on a patch. 

%We retrieved the equations (\ref{eq:memory_hist}) for computational complexity per frame ($C_{HIST}$) and memory requirement in bits ($M_{HIST}$) of HIST RP from~\cite{AuthorsEBBIOT}. 
The computational and memory complexity of HIST RP are reported in \cite{AuthorsEBBIOT}. The corresponding equations for CCL RP labeled as $C_{CCL}$ and $M_{CCL}$ as derived in eq. \ref{eq:memory_ccl}, depend on the parameter $\alpha$ since the main comparisons in the algorithm happen only on active pixels. The first term of $C_{CCL}$($M_{CCL}$) denotes the contribution of downsizing. The second term in $C_{CCL}$ denotes the computation only on active pixels. We can keep a fixed memory assuming that we have maximum number of equivalent labels, possible only when there is an inactive pixel between every two active pixels. Therefore, the second term in $M_{CCL}$ indicates the memory required for storing the four BB corners for each equivalent label.
% \begin{align}
% \label{eq:memory_hist}
%     C_{HIST} &=  A\times B + 2\frac{A\times B}{s_1s_2}\notag\\
%     M_{HIST} &= \frac{A\times B}{s_1s_2} \lceil log_2(s_1s_2)\rceil+ \notag\\ 
%      &(\frac{A}{s_1}\lceil log_2(B\times s_1) \rceil + \frac{B}{s_2}\lceil log_2(A\times s_2) \rceil)
% \end{align}
\begin{align}
\label{eq:memory_ccl}
    C_{CCL} &=  A\times B + \alpha\frac{A\times B}{s_1s_2}\notag\\
    M_{CCL} &= \frac{A\times B}{s_1s_2} + \notag\\
     &(\frac{A\times B}{2s_1s_2}\lceil log_2(\frac{A}{s_1})\rceil + \frac{A\times B}{2s_1s_2}\lceil log_2(\frac{B}{s_2}) \rceil)
\end{align}

For our specific case, we estimated $\alpha$ to be between $2.7$ and $4.5$, by running CCL RP over the dataset as discussed later in the paper. Combining that with the sensor dimensions for DAVIS camera $A=240$, $B=180$ and well fitting scaling factors $s_1=6$, $s_2=3$ for our case, we estimated that HIST RP performs $C_{HIST}=48$ Kop/frame and $M_{HIST}=3.44$ KB while CCL RP has maximum $C_{CCL}\approx54$ Kop/frame ($\alpha = 4.5$) and $M_{CCL}=16.8$ KB. Although, the number of computations are similar for both HIST and CCL RPs, the memory requirement increases five fold for CCL. However, it should be noted that such increase does not play much role in the system level since it is much less than the memory required by NNDC as shown in the following sub-section.

\subsubsection{Combining CCL and NNDC RP}
\label{sec:NNDCrp}

Although there are low-cost frame-based single step object detector and classifier solutions in the literature such as YOLO~\cite{redmon2016yolo9000,redmon2018yolov3}, SSD-MobileNet~\cite{Liu_2016}, in order to target for stand-alone IoVT devices based real-time traffic monitoring, implementing such CNN based networks in compact, power-constrained hardware ($<1$ mW) is not feasible.

CCL RP discussed earlier, plays a fundamental role in recognizing salient information from the achromatic binary image, but does not cover highly fragmented objects such as buses, trucks in some of the scenes generating more than two RPs for single objects (shown in Fig.~\ref{fig:rps_comparison}(b)). Therefore, a secondary correction step for removing unwanted RPs and merging BBs is required. However it will require the knowledge about the RP and its associated class in order to merge them~\cite{redmon2016yolo9000}. Keeping in mind the memory constraints, we propose a C\underline{NN} based \underline{D}etector (position correction) plus \underline{C}lassifier model (NNDC RP) which predicts the class and confidence for the RP, and correctly modifies the position of RP bounding box. 
% This model works frame-by-frame with no inter-association between frames at different timings. 
% Hence, with the motivation 

The initial inspiration for training this model came from YOLOv2~\cite{redmon2016yolo9000} wherein, the idea of predicting BB coordinates offsets and usage of hand-picked anchor boxes (priors) was proposed. We borrow these ideas from YOLOv2 and apply them to train a variant of LeNet5~\cite{lecun1995comparison,lecun1998gradient}, with a $42\times42\times2$ input, cropped from the centroid or symmetrically zero-padded image from RP bounding box coordinates of 1B2C frame. LeNet5 was chosen due to the similarity in the image sizes and the number of training images between our dataset and MNIST used to train LeNet5\cite{lecun1998gradient}. Variants of YOLOv2 were considered as a backbone but the huge difference between conventional RGB images and EBBI precluded us from using pre-trained networks. Also, the small size of the dataset ($\approx 10X$ less than ImageNet) prevented us to train such big networks from scratch.

The network produces $C + 5$ outputs including confidences for all available classes ($C$), objectness score BB\textsubscript{conf}) and bounding box correction parameters ($t_x, t_y, t_w, t_h$). This model differs from YOLOv2 in the following aspects: (a) in place of the entire frame, the input to the model is RP obtained from CCL, (b) the anchor boxes are determined from mean sizes of class categories each representing one of the classes, unlike k-means clustering used in YOLOv2, and (c) the prediction contains just one bounding box per input RP instead of multiple bounding boxes for each grid cell of the input frame. The rest -- hidden layers, activations, number of filters, filter sizes for convolution layers, in the modified model are kept the same, except for BB\textsubscript{conf} and BB correction parameters which have linear activation. Physically, BB\textsubscript{conf} represents whether the RP being analyzed contains sufficient information about the object or not, and a threshold $(\rm{thr})$ to it helps in flagging the RP for rejection or consideration for passing to tracker. BB correction parameters $(\hat{t}_x, \hat{t}_y)$ represent the predicted offset for upper left corner $\rm{(RP_x,RP_y)}$ of RP bounding box, while $(\hat{t}_w, \hat{t}_h)$ represent the predicted width and height correction parameters for the box's width and height $\rm{(RP_x,RP_y)}$. 

We note that predicting the offsets, $(\hat{t}_x, \hat{t}_y)$ is advantageous and makes training smoother~\cite{redmon2016yolo9000}. Intuitively, the same percentage error in offset prediction results in smaller overall position error than predicting the positions directly. However, learning the sizes of the objects is the most important aspect for the model and therefore, we feed the knowledge of priors to the model. We ensure that the number of priors are equal to the number of classes, $C$, with each prior corresponding to a class. The anchor box sizes are determined from the mean sizes of ground truth (GT) BBs for each of the classes in the input dataset. The new size of RP is predicted using the anchor box size of the predicted class and size correction parameters for the RP. The complete algorithm for the calculation of corrected RP location is shown in Algorithm \ref{alg:new_locs}.

% \label{eq:newPos}

\begin{algorithm}[t]
\SetAlgoLined
\SetKwInOut{KwIn}{Input}
\SetKwInOut{KwOut}{Output}

\KwIn{A list $[(w_i,h_i)]$, $i=1, 2, \cdots, C$, where each tuple is 
anchor box size for class $i$. \newline
A list $[\hat{o}_i]$, $i=1, 2, \cdots, C$, where each element is 
predicted confidence for class $i$.\newline
Bounding Box predicted correction parameters: [$\rm{\hat{BB}_{conf}}$$,\hat{t}_x, \hat{t}_y, \hat{t}_w, \hat{t}_h$]\newline
Initial location of RP's top left corner: $\rm{(RP_x,RP_y)}$ }
\KwOut{New Region Proposal BB Location: [$\hat{x},\hat{y},\hat{w},\hat{h}$]}
% \KwResult{Write here the result }
%  initialization\;
%  \While{While condition}{
%   instructions\;
% \begin{algorithmic}[1]
  \eIf{$\rm{BB_{conf} < thr}$}{
   Box is rejected\;
   }{
   find $j$, $\rm{max}$$(\hat{o}_j)$ where $j\in1, 2, \cdots, C$\;
   for that $j$, get $(w_j, h_j)$\;
   $\hat{x} = $$\rm{clip(tanh}$$(\hat{t}_x)*(A - 1) + $$\rm{RP_x}$$, 0, A - 1) $\;
   $\hat{y} = $$\rm{clip(tanh}$$(\hat{t}_y)*(B - 1) + $$\rm{RP_y}$$, 0, B - 1) $\;
   $\hat{w} = $$\rm{clip}$$(w_j*$$\rm{exp}$$(\hat{t}_w), 0, A)$\;
   $\hat{h} = $$\rm{clip}$$(h_j*$$\rm{exp}$$(\hat{t}_h), 0, B)$\;
   where, $\rm{clip}$$(a,m,n)$ means $a$ is clipped with $m$ as lower bound and $n$ as upper bound
  }
%  }
% \end{algorithmic}
 \caption{New Position Calculation}
 \label{alg:new_locs}
\end{algorithm}

\textbf{Model Training:} While training the model, we gather all the RPs from all the training videos frame by frame and resize them into a fixed size of $42\times42\times2$, either by zero padding keeping the RP in centre or cropping it from the centroid. The true positions for each of the RPs for a particular frame are defined according to the intersection-over-union, IoU (eq. \ref{eq:IoU}) with the ground truth (GT) bounding boxes for that frame. If the IoU of RP with GT box is greater than IoU\textsubscript{th}=0.1, the true BB\textsubscript{conf} for that RP is assigned the same value as IoU and GT bounding box, $[x,y,w,h]$ act as true location for the RP; otherwise, BB\textsubscript{conf} is kept 0.
\begin{equation}
\label{eq:IoU}
    \rm{IoU}=\frac{A_{\rm{Intersection}}}{A_{\rm{Union}}}
\end{equation}
where A\textsubscript{Intersection} is the area of intersection and A\textsubscript{Union} is the area of union of RP box and the GT box. Therefore, we form the new loss function (eq. \ref{eq:lossfn}) combining the three components given by:
% \textcolor{red}{
% \begin{align}
% \label{eq:lossfn}
%     &\rm{Loss_1} = \sum_{i=1}^{C}{(o_i-\hat{o}_i)^2}\notag\\
%     &\rm{Loss_2} = (BB_{conf}-\hat{BB}_{conf})^2\notag\\
%     &\rm{if}\, BB_{conf}\, >\, 0.1,\notag\\
%     &\rm{Loss_3} = (\frac{x-\hat{x}}{A-1})^2+(\frac{y-\hat{y}}{B-1})^2+(\frac{w-\hat{w}}{A})^2+(\frac{h-\hat{h}}{B})^2 \notag\\
%     &\rm{else},\notag\\
%     &\rm{Loss_3} = 0\notag\\
%     &\rm{Total\, Loss} = \rm{Loss_1} + \rm{Loss_2} + \lambda*\rm{Loss_3}
% \end{align}
% }
% \textcolor{red}{
\begin{align}
\label{eq:lossfn}
    &\rm{Loss_1} = \sum_{i=1}^{C}{(o_i-\hat{o}_i)^2}\notag\\
    &\rm{Loss_2} = (BB_{conf}-\hat{BB}_{conf})^2\notag\\
    &\rm{Loss_3} = \begin{cases}
  (\frac{x-\hat{x}}{A-1})^2+(\frac{y-\hat{y}}{B-1})^2+\notag\\
    \qquad (\frac{w-\hat{w}}{A})^2+(\frac{h-\hat{h}}{B})^2 \text{, if } BB_{conf} > 0.1 \\
  0 \text{, otherwise}
    \end{cases}\notag\\
    &\rm{Total\, Loss} = \rm{Loss_1} + \rm{Loss_2} + \lambda*\rm{Loss_3}
\end{align}
% }
where, Loss\textsubscript{1} is the classification error, Loss\textsubscript{2} is the bounding box confidence error and $\lambda$ is the Lagrange multiplier used to give appropriate weight to Loss\textsubscript{3}. It also helps the model to give attention to better position detection. This loss function is largely modified from YOLOv1~\cite{redmon2015look}, with the penalization for BB coordinates (Loss\textsubscript{3}) being changed according to the IoU of RP box with the GT box, and the width and height of boxes optimized directly instead of their square roots.

While testing the model, the predicted $\rm{\hat{BB}_{conf}}$ helps in rejecting the RPs and the new BB coordinates are predicted only if $\rm{\hat{BB}_{conf}}$ is greater than the assigned threshold, $\rm{IoU_{th}}$. Therefore, the knowledge of priors gives an upper-hand in predicting finely localized box and the corresponding class information. \\This object detector, however may be left with multiple overlapping boxes for the same object after prediction. Consequently, we suggest the application of three-step greedy non-maximal suppression (NMS)~\cite{hosang2017learning} for removing the unwanted overlapping boxes: 
\begin{itemize}
    \item Sort the new BBs for a particular frame according to the predicted $\rm{\hat{BB}_{conf}}$.
    \item Start with the best scoring box and find its IoU with the other BBs one-by-one and suppress the other BB if IoU is greater than a fixed threshold, $\rm{thr^{ns}}$.
    \item Repeat the same procedure with the next box in the sorted array until no extra boxes remain in the list.
\end{itemize}
\begin{figure}[t]
\centering     %%% not \center
\includegraphics[width=\linewidth]{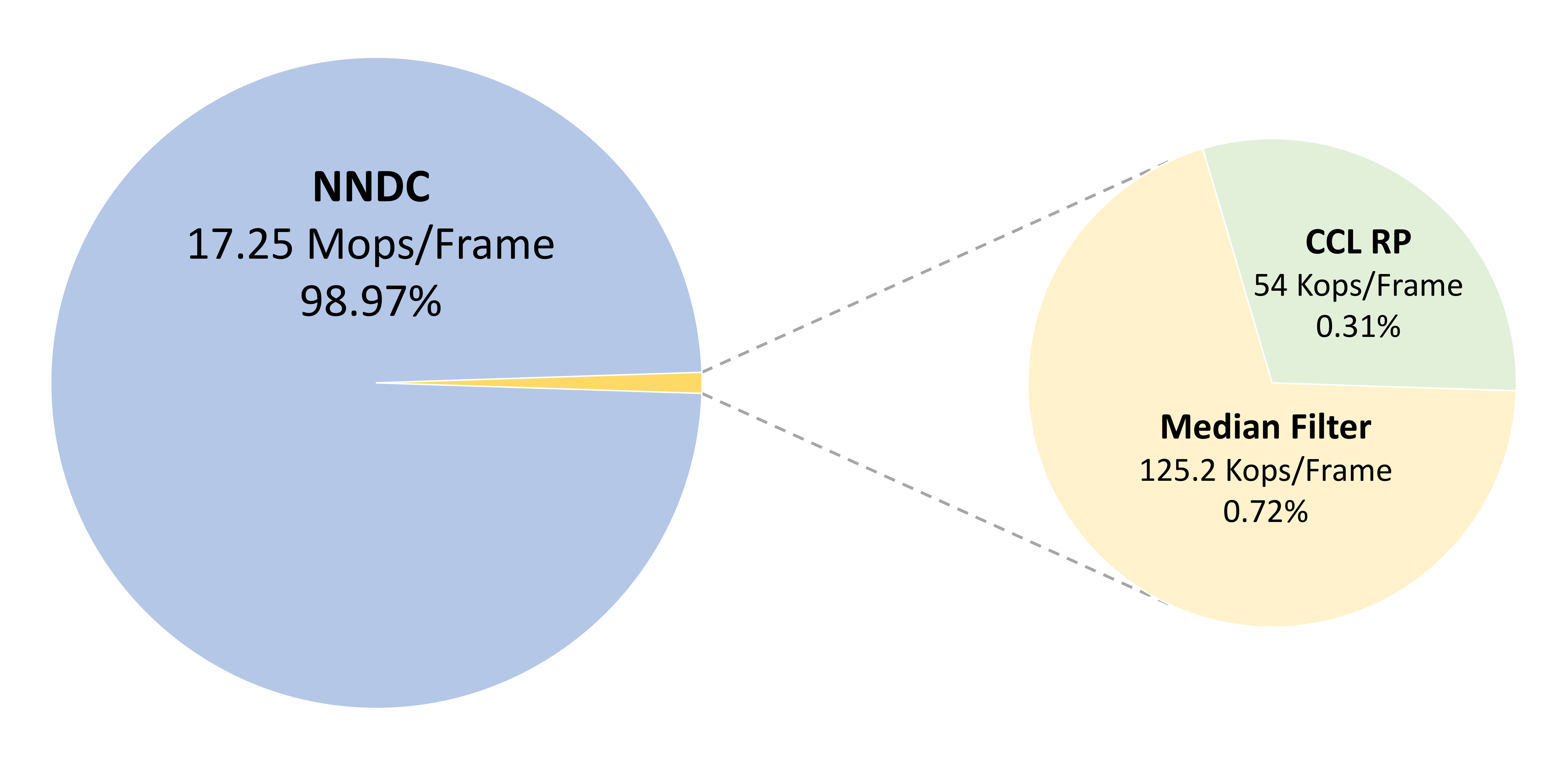}
 \caption{Computational Cost Distribution for EBBINN on active frames (not considering temporal sparsity) assuming $8$ regions proposed by CCL as input to the neural network.}
\label{fig:comp_cost_pie_chart}
\end{figure}
% It is to be pointed out that the formulation for new bounding box calculation may seem unconstrained, but in reality, it does not shift BBs much because the RPs already contain fragmented objects and the actual complete object is just at a small offset and has size as a factor of the prior class size.
It is to be pointed out that the calculation for rectified bounding box may seem unconstrained, however, the knowledge of RP being a part of the complete object allows the network to learn that the actual bounding box of required object is at a small offset of RP and has size as a factor of the prior class size. 

Further, this modified version of LeNet5 has much fewer computes than other object detector models (shown in Table~\ref{tab:computes}). For NNDC, the \emph{average} computations for a video, $\overline{C_{NNDC}}$, depend on the temporal occupancy of frames, $\alpha_T$, the average number of RPs, $\overline{n_{\rm{RP}}}$, and can be expressed as shown below in eq. \ref{eq:comp_nndc}:

\begin{equation}
\label{eq:comp_nndc}
    \overline{C_{NNDC}}=\alpha_T \overline{n_{\rm{RP}}}C_{\rm{NNDC}}
\end{equation}

where the number of computations multiplications and additions) for running the neural network oncefor a single RP is $C_{\rm{NNDC}}=2.16$M. Since $\alpha_T$ depends on the video, we keep its value as $1$ here but include actual values from data in Section \ref{sec:mod_results}. Also, we choose a worst case value of $n_{\rm{RP}}=8$ and for a fair comparison with other models, we combine the computes for EBBI and CCL RP leading to a total computes bound per frame of $\approx17.302M$ for our proposed approach as shown in Fig.~\ref{fig:comp_cost_pie_chart}. Note that these two blocks do not add much computes to the overall total, showing that most of the computation is done in NNDC. Nonetheless, one of the largest factor contributing to the smaller operations for NNDC is the smaller input size arising from simplified RP process. It can be seen that Tiny YOLOv2, YOLOLite and SSD-MobileNet have $\approx52\times, \approx24\times, \approx16.8\times$ higher computes per frame respectively than this model.  The computes and parameters for other models were calculated on an image size of $240\times180\times2$, which is the sensor dimensions of DAVIS240C with information in ON-OFF polarity channels.  

\begin{table}[t]
\caption{Computations for different object detector and classifier models}
\centering
%\begin{tabular}{|r|r|r|r|r|r|r|}
\begin{tabular}{lll}
\hline
{\color[HTML]{000000} \textbf{Network}} & {\color[HTML]{000000} \textbf{Total \# Computes}} & 
{\color[HTML]{000000} \textbf{\# Parameters}} \\ \hline
{\color[HTML]{000000} \textbf{NNDC}} & {\color[HTML]{000000} \textbf{2.16-17.3M}} & {\color[HTML]{000000} \textbf{0.108M}} \\
{\color[HTML]{000000} Tiny YOLOv2~\cite{redmon2016yolo9000}} & {\color[HTML]{000000} 898M} & {\color[HTML]{000000} 15.74M} \\
{\color[HTML]{000000} YOLOLite~\cite{pedoeem2018yololite}} & {\color[HTML]{000000} 418M} & {\color[HTML]{000000} 0.542M} \\
{\color[HTML]{000000} SSD-MobileNet~\cite{Liu_2016}} & {\color[HTML]{000000} 290M} & {\color[HTML]{000000} 26.34M}\\ \hline
%{\color[HTML]{000000} SiamMask~\cite{DBLP:Siammask}} & {\color[HTML]{000000} $\approx38000M$} & {\color[HTML]{000000} $\approx157M$}\\ \hline
\end{tabular}
\label{tab:computes}
\end{table}
%------------------------------

\subsection{Overlap based Tracking}
\label{sec:OT_subsection}
Inspired by KF which works on the principle of prediction and correction, we present a simpler tracker that takes advantage of three properties of stationary DVS: (a) rejection of background, (b) possibility of high frame rates ($\approx 200$ Hz) and (c) low event rates. Due to these three factors, the assignment of detections to tracks can be simplified to just checking overlap followed by greedy assignment, hence the name overlap based tracker (OT). Occlusion is handled by having extra checks based on predicted trajectories, assuming a constant velocity model. OT works on the principle of prediction of current tracker position from past measurements and correction based on inputs from the region proposal (RP) network~\cite{AuthorsEBBIOT}.One assumption for the correct operation is when an object enters the scene, it is not immediately occluded. If this is violated, the corresponding track cannot be corrected. Also, as mentioned in Section \ref{subsec:ebframegen}, $t_F=66$ ms equivalent to a frame rate of $\approx 15$ Hz was found sufficient for our application to guarantee significant overlap between objects in consecutive frames.
Using $P_i$ and $T_i$ ($1\leq i\leq 8$) to represent bounding boxes obtained from the region proposal network and OT respectively, each composed of upper-left corner coordinates (x,y) and object dimensions (w,h),  the major steps performed by the OT for each EBBI frame, can be summarized as follows:
\begin{enumerate}
    \item The  tracker is initialized and the predicted position $T_i^{pred}(x,y)$ of all valid trackers is obtained by adding $T_i(x,y)$ with corresponding horizontal ($V_x$) and vertical ($V_y$) velocity. This is equivalent to a constant velocity assumption between successive frames and is reasonable given the high frame rates of DVS as well as the lack of camera motion.
    \item For each valid tracker $i$ in the \emph{tracking} or \emph{locked} mode, $T_i^{pred}$ is matched with all available region proposals $P_j$. A match is found if overlapping area between the $T_i^{pred}$ and $P_j$ is larger than a certain fraction of area of the two ($T_{ov}$) i.e., $overlap(T_i^{pred},P_j)>T_{ov}\implies Match Found$  -- hence the name overlap based tracker (OT).
    \item If a region proposal $P_j$ does not match any existing tracker and there are available free trackers, then a new tracker $T_k$ is seeded and initialized with $T_k=P_j$. Every new tracker is initially set to \emph{tracking mode} with no track count assigned to it. Once the new tracker matches one or more region proposals, it is set to \emph{locked mode} and a track count is assigned to it.
    \item If a $T_i^{pred}$ matches single or multiple $P_j$, assign all $P_j$ to it and update $T_i$ and velocities as a weighted average of prediction and region proposal. Here, past history of tracker is used to remove \emph{fragmentation} in current region proposal if multiple $P_j$ had matched.
    \item A $P_j$ matching multiple $T_i^{pred}$, can be a result of two possible scenarios--first, due to dynamic occlusion between two moving objects and second, assignment of multiple trackers to an object resulting due to region proposals corresponding to a fragmented object in the past. An occlusion is detected if the predicted trajectory of those trackers for $n=2$ future time steps result in overlap. For tracker undergoing occlusion, $T_i$ is updated entirely based on $T_i^{pred}$ and previous velocities are retained. In the case of multiple matching trackers resulting from an earlier region proposal of a fragmented object, the multiple $T_i^{pred}$ are merged into one tracker based on $P_j$ and corresponding velocity is updated. The other trackers are freed up for future use.
\end{enumerate}
The computational complexity of the OT algorithm is described in Sec. 3 of the \emph{Supplementary Material} \cite{ebbinnot-supplement}. Numerical evaluation of this equation and comparison with KF will be done later in Section \ref{subsec:Computational_Cost}. At the system level, these are also overshadowed by the NNDC computations.
\subsection{Tracker Class Assignment}
Our work in EBBIOT~\cite{AuthorsEBBIOT} does not have a mechanism for assigning classes to a tracker, $T_i$. However, with the outputs of NNDC RP acting as input to the tracker, we resolve the problem of class assignment to the detected trackers based on the following criteria:
\begin{itemize}
    \item If the number of matched RPs to the tracker, $T_i$ is one, assign the same class to $T_i$.
    \item Otherwise, if more than one RPs are matched to $T_i$, select the class with highest class confidence in the combined list of class confidences of all the matched RPs. This is the new assigned class to $T_i$.
    \item If dynamic occlusion between two tracks is detected in the frame, the class assignment is stopped for both of them and these track points are not considered for voting of class for the whole track.
\end{itemize}

To summarize, the event information from DVS goes into EBBI block generating 1B1C and 1B2C images. After application of median filtering on 1B1C image, it is sent to CCL RP and then, the generated RPs are further passed to NNDC block in the form of $42\times42\times2$ images containing 1B2C image of the object. The new modified RPs from NNDC are further passed to the OT for generating the trackers along with their classification. The next section will showcase the results for the described methodology.

\section{Results}
\label{sec:mod_results}
This section presents the data collection process followed by the evaluation of the proposed noise filtering technique. Then, we show the training of our classification model and provide insights about the hybrid RP network for the pipeline along with its comparison to other RP networks. Next, we compare the OT and KF trackers, followed by comparing the full EBBINNOT pipeline with event-based and frame-based state-of-the-art methods in Section \ref{sec:statecompare}. Finally, we compare the computations and memory usage of proposed EBBINNOT with other methods. The different pipelines studied and presented in this work were implemented in MATLAB. The CNN in the NNDC RP was trained on an NVIDIA TITANX GPU using a Python Keras implementation and the saved weights loaded into MATLAB using the Deep Learning Toolbox for model testing.
\begin{figure}
\centering
\subfloat[Site 1] {{\includegraphics[width=7cm]{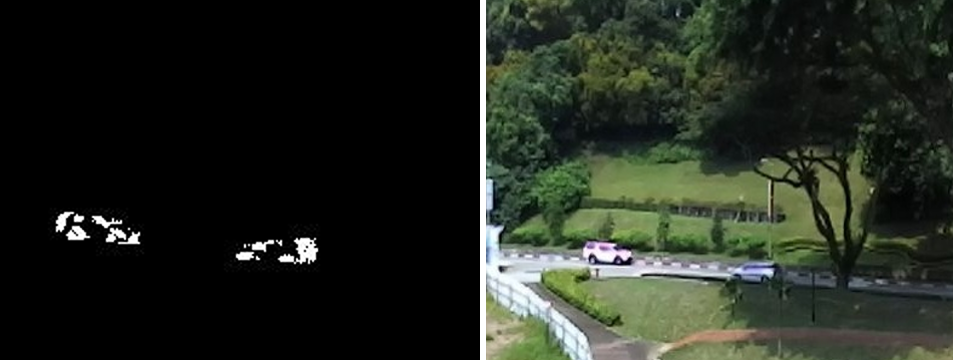}}\label{ENG_loc}}
\vspace{0.1em}
\subfloat[Site 2] {{\includegraphics[width=7cm]{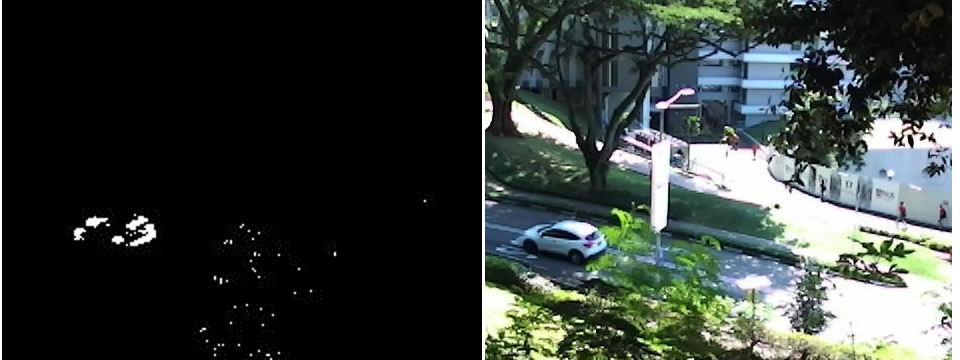}}\label{LT4_loc}}
\vspace{0.1em}
% \subfloat[Site  3] {{\includegraphics[width=9cm]{img/site_locations/YIH.png}}}
% \vspace{0.1em}
\subfloat[Site 3] {{\includegraphics[width=7cm]{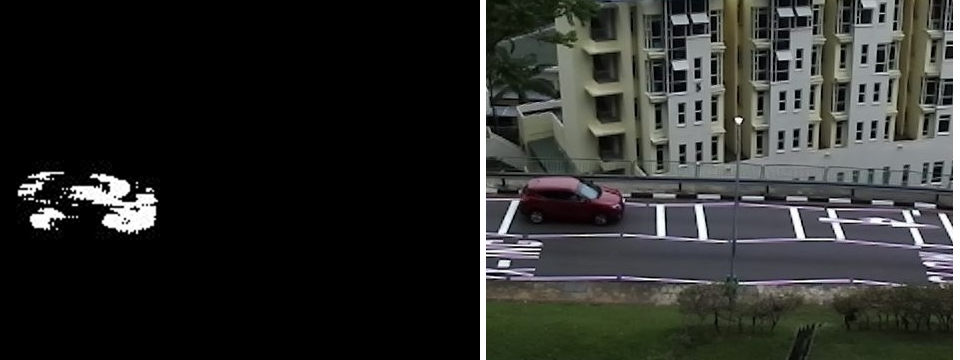}}\label{PGP_loc}}
\caption{Visual representation of datasets, i.e EBBI (Left) and RGB Image (Right), recorded at various sites discussed in Table \ref{tab:dataset_statistics_overview}}
\label{fig:Locations}
\end{figure}
\begin{figure}[t]
\centering     %%% not \center
\includegraphics[width=70mm]{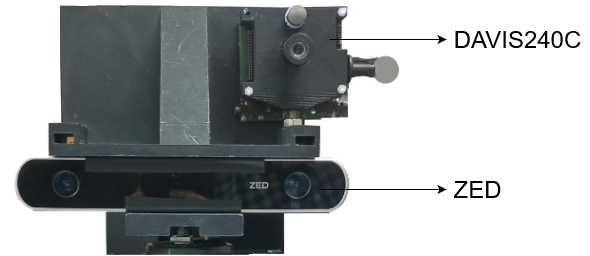}
 \caption{Experimental setup with ZED (Focal length : $2.8 mm$, Aperture: f/2.0) \cite{ZED} and DAVIS240C (Focal Length: Refer to column 4 of Table \ref{tab:dataset_statistics_overview}) \cite{brandli2014240} mounted on custom-made 3D mount}
\label{fig:Dataset_collection_setup}
\end{figure}

\subsection{Data Acquisition}
\begin{figure*}[t]
%%% not \center
\resizebox{\textwidth}{!}{
\begin{tabular}{ccc}
\subfloat[]{\includegraphics[width=.3\textwidth]{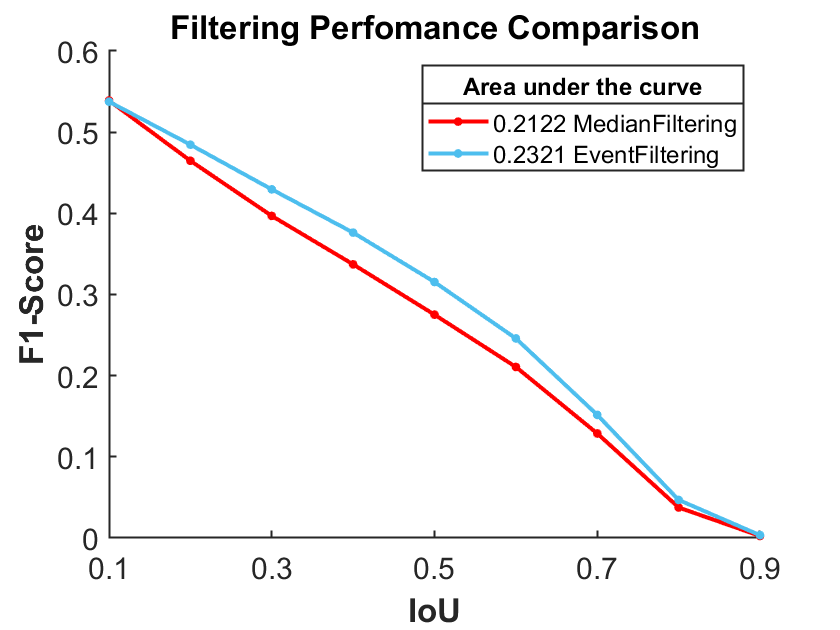}} &
\subfloat[]{\includegraphics[width=.3\textwidth]{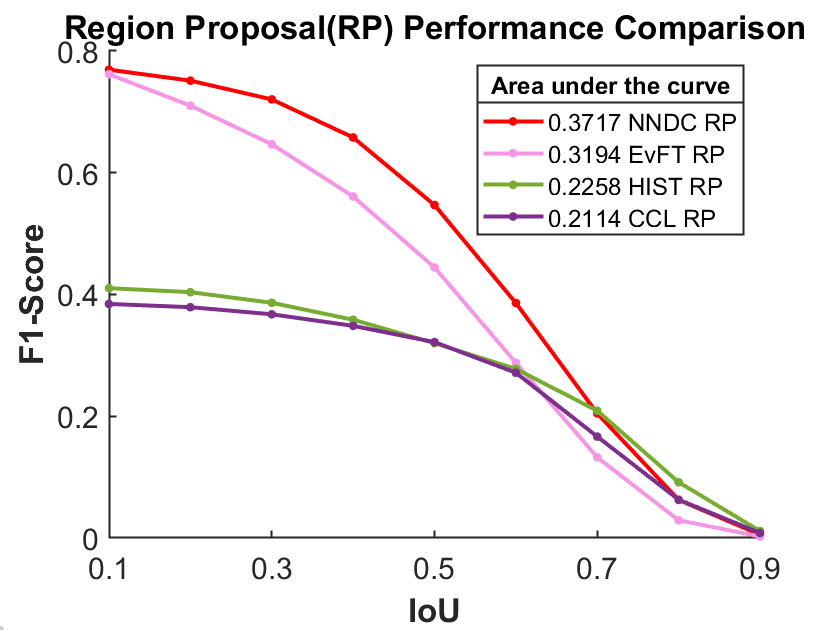}} &
%%\subfloat[]{\includegraphics[width=.33\textwidth]{img/plots/rpComparison_r.png}} &
\subfloat[]{\includegraphics[width=.3\textwidth]{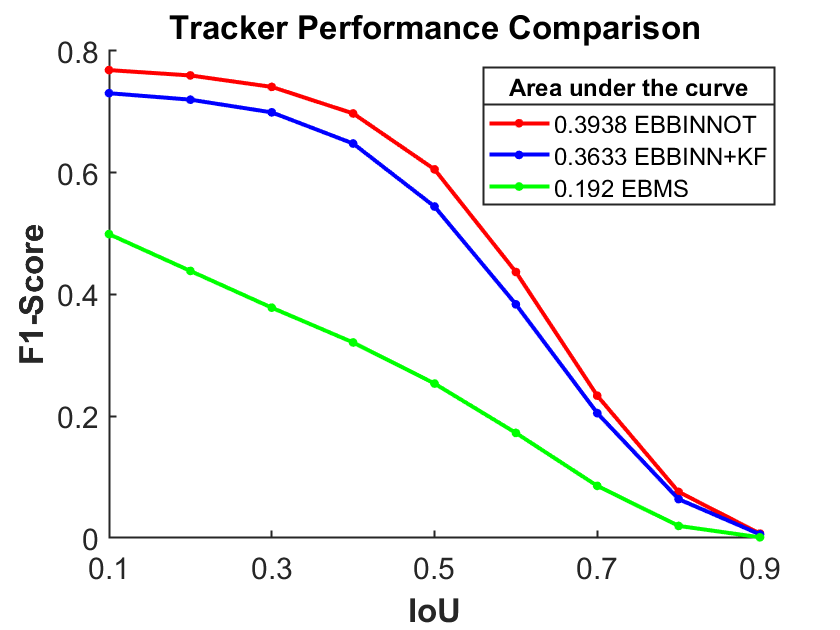}}
\end{tabular}
}
 \caption{Detection performance improves with each block of the pipeline, (a) Filtering Effect [Sec. \ref{subsec:Noise_Filtering_Comparison}]: Event-based filtering (Refractory Filter + NN-Filter + EBBI) Vs. Median Filtering (EBBI + Median Filter) followed by CCL RP showing comparable F1-scores with event-based filtering slightly performing superior. (b) RP Performance [Sec. \ref{subsec:rpn_comparisons}]: NNDC (CCL+NNDC) performs much superior than others in a EBBI+RP setup. (c) Tracker Performance [Sec. \ref{sec:flowcompare}]: Comparison between trackers show EBBINNOT to be best in terms of weighted F1-Score.  [List of Abbreviations is available on page 1]
 \\
  } 
\label{fig:filtering_RP_track_Comparison}
\end{figure*}
In this paper, we wanted to compare performance of DVS with a standard RGB camera; however, such a dataset is not available as far as we know. Consequently, it demanded the acquisition of event-based data and RGB data from a real traffic scenario for training, validation and testing \footnote{Dataset: \url{https://nusneuromorphic.github.io/dataset/index.html}}.

The chosen location for the traffic recordings was a high, perpendicular view from the road near intersections. In this regard, three places shown in Figure~\ref{fig:Locations}, were chosen for data collection using DAVIS240C. Further, we also captured RGB recordings for simultaneous comparison with the purely frame-based tracking SiamMask \cite{DBLP:Siammask} \& SiamRPN++ \cite{SiamRPN++_Li_2019_CVPR}, as shown in Figure \ref{fig:Dataset_collection_setup}. 
%For both the RGB and event datasets, manual ground truth (GT) annotation was carried out to facilitate tracker and classifier evaluation.
We manually labelled all of recorded RGB and event datasets using a custom MATLAB code. The RP and class labels were annotated for every 66ms of the recordings. In addition, the event and RGB data were made to have similar field-of-view (FoV) for a close comparison.

\subsection{Evaluation Metrics}
In order to test the system performance, we employed two evaluation metrics for object detection, classification and tracking.
\begin{itemize}
    \item \textbf{F1 score for detection performance:} We have already discussed in Section~\ref{sec:NNDCrp} that IoU is an effective metric for evaluating the detection accuracy. The tracker annotation can be matched with GT annotation to get IoU in order to conclude whether it represents a true object BB ($\rm{IoU>IoU_{th}}$) or a false object BB according to the threshold ($\rm{IoU_{th}}$). Thereafter, we sweep $\rm{IoU_{th}}$ from $0.1-0.9$ in steps of $0.1$ to find out precision and recall averaged over the entire duration of the recording. We further calculate F1 score (eq. \ref{eq:f1_score}) at each $\rm{IoU_{th}}$ as follows:
    \begin{align}
    \label{eq:f1_score}
    &F1^j=2\frac{P^j\times R^j}{P^j+ R^j}\notag\\
    &F1^{\rm{wtd}} = \frac{\sum_{j=1}^{K}N^j\times F1^j}{\sum_{j=1}^{K}N^j}
    \end{align}
    Here, $P^j$ and $R^j$ are precision and recall for the recording $j$, $N^j$ represents number of tracks in recording $j$ and $F1^{\rm{wtd}}$ represents the weighted F1 score for all the $K$ recordings, $j=1,\dots, K$. Thus, we examine the detection performance of our dataset in terms of $F1^{\rm{wtd}}$ swept over $\rm{IoU_{th}}$.
    \item \textbf{Overall accuracies for classification performance:} We calculated per-sample and per-track classification accuracies. In order to calculate the predicted class of a track, we recorded the statistical mode of the classification output for all the samples in the respective track of a vehicle. Further, we defined two types of accuracies: overall balanced and overall unbalanced. The former represents the average of class-wise accuracies to have a definitive evaluation measure while dealing with the dataset imbalance. The latter represents the widely used average accuracy for all the samples in the dataset regardless of class distribution. 
\end{itemize}

\begin{table*}
\caption{DAVIS240C traffic dataset}
\centering
\resizebox{\textwidth}{!}{%
\begin{tabular}{cccccccccc}
\hline
{\color[HTML]{000000} \textbf{\begin{tabular}[c]{@{}c@{}}Recording \\ Site\end{tabular}}} & {\color[HTML]{000000} \textbf{Duration}} & {\color[HTML]{000000} \textbf{\begin{tabular}[c]{@{}c@{}}Time \\ of Day\end{tabular}}} & \textbf{\begin{tabular}[c]{@{}c@{}}Lens \\ Focal Length\end{tabular}} & {\color[HTML]{000000} \textbf{\begin{tabular}[c]{@{}c@{}}Number of \\ Events\end{tabular}}} & {\color[HTML]{000000} \textbf{\begin{tabular}[c]{@{}c@{}}\# of Recordings\\ in Training $\mid$ Testing\end{tabular}}} & {\color[HTML]{000000} \textbf{\begin{tabular}[c]{@{}c@{}}Car/Van  \\ (Samples $\mid$ \\ Tracks)\end{tabular}}} & {\color[HTML]{000000} \textbf{\begin{tabular}[c]{@{}c@{}}Bus  \\ (Samples $\mid$ \\ Tracks)\end{tabular}}} & {\color[HTML]{000000} \textbf{\begin{tabular}[c]{@{}c@{}}Bike  \\ (Samples $\mid$ \\ Tracks)\end{tabular}}} & {\color[HTML]{000000} \textbf{\begin{tabular}[c]{@{}c@{}}Truck  \\ (Samples $\mid$ \\ Tracks)\end{tabular}}} \\ \hline
{\color[HTML]{000000} \textbf{Site 1}} & {\color[HTML]{000000} 2h11m} & {\color[HTML]{000000} 3PM, 4PM} & 12mm & {\color[HTML]{000000} 201M} & {\color[HTML]{000000} 6 $\mid$ 2} & {\color[HTML]{000000} 18232 $\mid$ 379} & {\color[HTML]{000000} 8081 $\mid$ 165} & {\color[HTML]{000000} 1378 $\mid$ 35} & {\color[HTML]{000000} 2256 $\mid$ 47} \\
{\color[HTML]{000000} \textbf{Site 2}} & {\color[HTML]{000000} 2h25m} & {\color[HTML]{000000} 3PM, 4PM} & 6mm & {\color[HTML]{000000} 132M} & {\color[HTML]{000000} 6 $\mid$ 3} & {\color[HTML]{000000} 16918 $\mid$ 382} & {\color[HTML]{000000} 8019 $\mid$ 177} & {\color[HTML]{000000} 1604 $\mid$ 39} & {\color[HTML]{000000} 2513 $\mid$ 56} \\
{\color[HTML]{000000} \textbf{Site 3}} & {\color[HTML]{000000} 1h} & {\color[HTML]{000000} 3PM} & 8mm & {\color[HTML]{000000} 50M} & {\color[HTML]{000000} 2 $\mid$ 1} & {\color[HTML]{000000} 6514 $\mid$ 209} & {\color[HTML]{000000} 1201 $\mid$ 27} & {\color[HTML]{000000} 512 $\mid$ 22} & {\color[HTML]{000000} 501 $\mid$ 15} \\ \hline
\end{tabular}%
}
\label{tab:dataset_statistics_overview}
\end{table*}

\subsection{Median filtered EBBI vs. Event-based noise filtering}
\label{subsec:Noise_Filtering_Comparison}
To evaluate the effect of the proposed median filtering approach on the detection performance of the whole pipeline, we replaced it with the commonly used AER event-based nearest neighbour filtering approach \emph{aka} Background Activity Filter (BAF) \cite{BAF_tobi,fnins2018, RameshFrontiers, Ramesh2019}. A correlation time of 5ms was found to result in better noise filtering for BAF on our datasets, after sweeping over a range of correlation time intervals of 0.5 to 10 ms. Therefore, a correlation time of 5 ms was used for the BAF and a window of $3\times3$ was used for the proposed median filtering approach.
%For a fair comparison, a correlation time of $5ms$ in a neighbourhood of $3\times3$ was implemented for the event filtering approach, similar to the $3\times3$ window used for the proposed median filtering approach. 
Since our proposed median filter with EBBI gives on par performance with the event-based filtering approach, as shown in Figure~\ref{fig:filtering_RP_track_Comparison}(a), we advocate it for low-power hardware implementations as carried out in this work. 

\begin{table}[t]
\caption{Mean object sizes at different recording sites}
\centering
\begin{tabular}{ccccc}
\hline
\textbf{Recording Site} & \textbf{Car/Van} & \textbf{Bus} & \textbf{Bike} & \textbf{Truck} \\ \hline
\textbf{Site 1} & $16\times42$ & $31\times94$ & $15\times21$ & $22\times50$ \\
\textbf{Site 2} & $25\times47$ & $52\times107$ & $17\times22$ & $35\times61$ \\
\textbf{Site 3} & $34\times82$ & $64\times180$ & $26\times44$ & $50\times104$ \\ \hline
\end{tabular}
\label{tab:class_sizes}
\end{table}
\subsection{Comparison of Region Proposal Networks}% is this heading correct
\label{subsec:rpn_comparisons}
\textbf{Data Preparation for NNDC training: }As mentioned in Section~\ref{subsec:ebframegen}, events were aggregated at a frame rate of $15$ Hz ($t_F=66$ ms) to form 1B1C and 1B2C frames. We noted that the size of objects played a significant role for the NNDC model since an anchor box guides the class size. The objects at site 3 location had significantly different mean class sizes when compared to other sites (shown in Table~\ref{tab:class_sizes}). Therefore, to facilitate the model training, we rescaled the frame by half to $120\times90$ at site 3 location using nearest neighbor interpolation.

Table~\ref{tab:dataset_statistics_overview} shows the statistical distribution of the dataset in terms of the number of samples obtained for each class category, and the number of recordings kept from each site for training and testing. The $42\times42\times2$ samples from the frames are obtained after applying CCL RP along with their correct positions, $\rm{BB_{conf}}$ and class information, by matching the respective samples with interpolated GT annotations. We also randomly selected $\approx63,000$ noisy samples obtained from CCL RP that did not match with any GT annotations (with $\rm{IoU<0.1}$) so that the network could classify them as a separate background class and give a predicted $\rm{\hat{BB}_{conf}}$ to each less than $\rm{thr=0.1}$. Assigning a different class was also necessary because these samples do not fit in any class category and in this class's absence, $\rm{Loss_1}$ could not be optimized.

Note that we did not consider samples from pedestrians in the training data acquisition since they generate very few events due to their small size and slow speed. Simultaneous tracking of pedestrians and vehicles is kept as a future work. In total, we had $C=5$ with classes: background, car/van, bus, bike and truck in our model with a total of $C+5 = 10$ outputs. Since the buses and trucks were generally bigger than the size of $42\times42$, we also included cropped samples from top-left, top-right, bottom-left and bottom-right sections of their RPs. This helped to reduce the class-wise sample variance, and also provided the information from the object's frontal and posterior region for tuned BB prediction. The bikes were augmented by random rotation within $\pm15^\circ$ and translation by some random amounts within the fixed area of $42\times42$. The samples from recordings assigned for testing were also collected using the same criteria, but without any noisy samples having an $\rm{IoU<0.1}$. The main objective of the training was to improve the $\rm{BB_{conf}}$, increase the BB actual overlap with the object, and also report its correct class.

\textbf{Training Details: }NNDC model was trained on $80\%$ of the training data randomly selected, while the rest was kept for validation. The model was trained on an NVIDIA TITANX GPU in the form of randomly shuffled batches of $128$ with $20$ assigned epochs, a learning rate of $0.01$ and $\lambda=5$ (hyperparameters optimized using grid search on validation set). This model, trained using Adam optimizer with default hyperparameters, was written in Keras framework because of the ease of writing custom loss functions like equation~\ref{eq:lossfn}. Further, the overall unbalanced accuracy metrics on validation data after each epoch were used for early stopping of the training with patience $3$. The best model was saved for evaluation on the test recordings collected at different times. 

\begin{table}[t]
\caption{Classification accuracies for testing samples recorded using DAVIS240C}
\centering

\begin{tabular}{lcc}
\hline
{\color[HTML]{000000} \textbf{Category}} & {\color[HTML]{000000} \textbf{per sample (\%)}} & {\color[HTML]{000000} \textbf{per track (\%)}} \\ \hline
{\color[HTML]{000000} Car/Van} & {\color[HTML]{000000} 86.59} & {\color[HTML]{000000} 95.8} \\
{\color[HTML]{000000} Bus} & {\color[HTML]{000000} 89.81} & {\color[HTML]{000000} 98.1} \\
{\color[HTML]{000000} Bike} & {\color[HTML]{000000} 81.02} & {\color[HTML]{000000} 100} \\
{\color[HTML]{000000} Truck} & {\color[HTML]{000000} 53.39} & {\color[HTML]{000000} 76.92} \\ \hline
{\color[HTML]{000000} \textbf{Unbalanced accuracy}} & {\color[HTML]{000000} \textbf{85.07}} & {\color[HTML]{000000} \textbf{95.39}} \\
{\color[HTML]{000000} \textbf{Balanced accuracy}} & {\color[HTML]{000000} \textbf{77.70}} & {\color[HTML]{000000} \textbf{92.70}} \\ \hline
\end{tabular}
\label{tab:DAVIS_scores}
\end{table}
\textbf{Inference: }Table~\ref{tab:DAVIS_scores} shows the per-sample as well as per-track accuracies on all the test recordings, including overall balanced and unbalanced accuracies. As expected, per track accuracies are higher due to the majority voting, and in the case of the Bike category, it is possible to get $100\%$ classification performance. We attribute this  to the unique size and shape of the bikes relative to the other categories. Overall, the balanced accuracy closely trails the unbalanced accuracy, which implies the classifier makes sound judgements instead of skewed decisions caused by the unbalanced DAVIS240C dataset. Different hardware friendly variants of the architecture have been tried for classification \cite{singla2020hynna} while maintaining the same depth. Separately, we have experimented  with increasing the depth and find $\approx 1\%$ accuracy improvement by adding one more layer, either convolutional or fully connected. Beyond this, there did not seem to be much improvement by increasing depth. However, we do not preclude the possibility of further improvement by exploiting latest advances in deep neural networks and keep it as a future work.

\textbf{EvFT RP Setup: }We implemented an event feature tracker based region proposal using the feature tracker described in \cite{Kostas-flow}, hereby referred to as EvFT RP, so that we can compare its performance with the other region proposal schemes discussed in this work. As a pre-processing step event-based NN-Filter with a period of 5ms was used for denoising the event data. Region proposals were generated using the  point cloud clustering algorithm \emph{pcsegdist()} in MATLAB by using the coordinates of the features as well as flow information obtained from the feature tracker. In other words, features that are close together and having similar flow are grouped together in a cluster. Rectangles enclosing the clusters are used as region proposals. 

\textbf{Overall RP Comparison: }In order to pick the best region proposal for the proposed pipeline, we ran the three RPNs, namely HIST, CCL and CCL + NNDC RP on the test dataset while restricting the maximum RPs to eight per frame. In this evaluation, the greedy NMS in NNDC had $\rm{thr^{ns}=0.3}$ for suppressing the boxes. Changing this parameter in the range of $0.2-0.4$ did not have appreciable effect on the results. To compare the performance at different IoUs, we used ground truth annotations at the same timestamps corresponding to the RPs. 
\par
Figure~\ref{fig:filtering_RP_track_Comparison}(b) shows the weighted F1 scores for the different RPs. Overall, the proposed CCL+NNDC RP significantly outperforms other RPs, as shown in Figure~\ref{fig:filtering_RP_track_Comparison}(b) with higher area under curve (AUC), calculated using trapezoidal numerical integration. Interestingly, HIST RP performs better than CCL RP by itself, due to lesser fragmentation by merging of overlapping regions. Integrating NNDC after CCL significantly improves this performance. Therefore, we adapt CCL+NNDC RP as part of our proposed pipeline and is also referred as hybrid RP. The closest performance is achieved by the EvFT RP generated from event-based flow computation. However, this comes at a much higher computational cost as shown later in Section \ref{subsec:Computational_Cost}. It should be noted however that EvFT RP would likely perform better for situations when the camera is moving as depicted in Figure \ref{fig:motivation}.
%\par It can also be seen that the proposed CCL+NNDC RP has slightly better accuracy than the EvFT RP. It is worthwhile to note that the number of computations in EvFT RP is also much higher than NNDC. Most computations in the EvFT RP (over $90\%$) are consumed by two EM flows of the feature tracker. 
%The RP creation process adds negligible overhead to the tracking process. 
%We estimated the number of computations required to implement the tracker + RP to be of the order of $1$ Mops per call of the EM functions by theoretical calculation and experimental verification by profiling the MATLAB code provided in \cite{Kostas-flow}. For a given recording in our dataset comprising $\approx 17$M events spanning $\approx16$ minutes,  the average number of computes per frame for  EvFT is $\approx11.5$ Mops, \textcolor{red}{which is YY times more than that required by EBBINN}.  

\begin{figure*}[t]
\centering
\resizebox{\textwidth}{!}{
\includegraphics[]{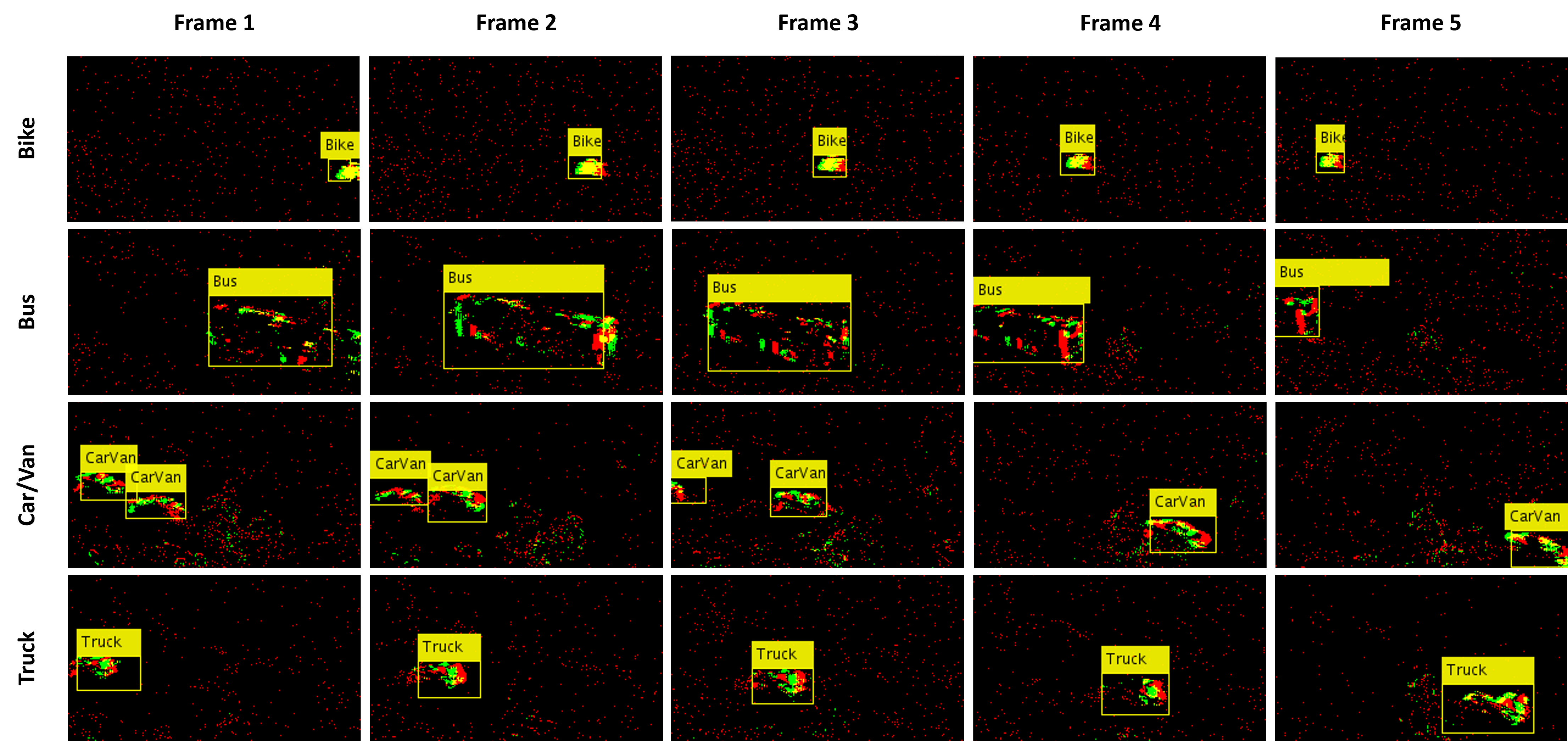}
}
\caption{Exemplar detection and classification results from the described EBBINNOT pipeline for tracks of different vehicles}
\label{fig:exemplar_tracks}
\end{figure*}
\subsection{Comparison of Tracker}
\label{sec:flowcompare}
For the purpose of fair comparison of performance of different trackers, we ensured that the same region proposal network, tracker parameters, tracker log generation method and evaluation metrics were used. For comparison of KF-Tracker and OT, the number of region proposals and trackers per frame were restricted to a maximum of $8$, the threshold for treating an object to be lost during tracking was set at invisibility for $5$ consecutive frames or less than $60\%$ visibility when the track is still valid. While for EBMS \cite{delbruck2013robotic}, the events were filtered using a refractory layer with accumulation period of $50$ ms, followed by NN-Filter with correlation time of $5$ ms. The minimum number of events required for cluster formation were kept $8$, maximum radius of cluster was kept $130$, and a time limit of $100$ ms was assigned in case of inactivity of cluster. These hyperparameter values were obtained after series of runs for optimization of EBMS on a set of short videos for validation. Figure \ref{fig:exemplar_tracks} illustrates the sample tracks generated for different types of vehicles for the trained EBBINNOT pipeline. Based on the observations made in~\cite{Authorsbmvc19}, we excluded tracks for human class while calculating the F1-scores for all the $5$ test dataset recordings excluding site 3.

As shown in Figure \ref{fig:filtering_RP_track_Comparison}(c), it can be noted that OT performs slightly better than KF and significantly better than the purely event-based EBMS tracker.
%\textcolor{red}{Need critical discussion on why -- for example, you did the ablation experiments by seeing which part of the OT code is responsible for most improvement} 
In order to ascertain the reason for performance improvement in OT as compared to KF, we performed an ablation study by removing specific parts of heuristics used in OT. Based on these comparisons, we can attribute the enhanced performance of OT to two reasons: first, the presence of a \emph{tracking} mode before transitioning to \emph{locked} state and second, the fragmentation handling logic in the OT. In our algorithm, only trackers in the \emph{locked} state are considered as a valid track. In the KF tracker with no \emph{tracking} mode, we observed that noisy event occurring intermittently results in false RPs creating new tracks for each of these noisy objects and increasing the false positives. As for fragmentation handling, unlike KF which cannot handle multiple trackers resulting from a fragmented object, OT utilizes past history of trackers to resolve a fragmentation in case multiple RPs match a tracker and merges multiple trackers that might be corresponding to an earlier fragmented RP, following the steps listed in Section \ref{sec:OT_subsection}. This logic effectively reduces multiple tracks being assigned to the same object and thereby boosts the performance of OT. 

\begin{figure}[t]
\scriptsize
\centering
\includegraphics[width=7cm]{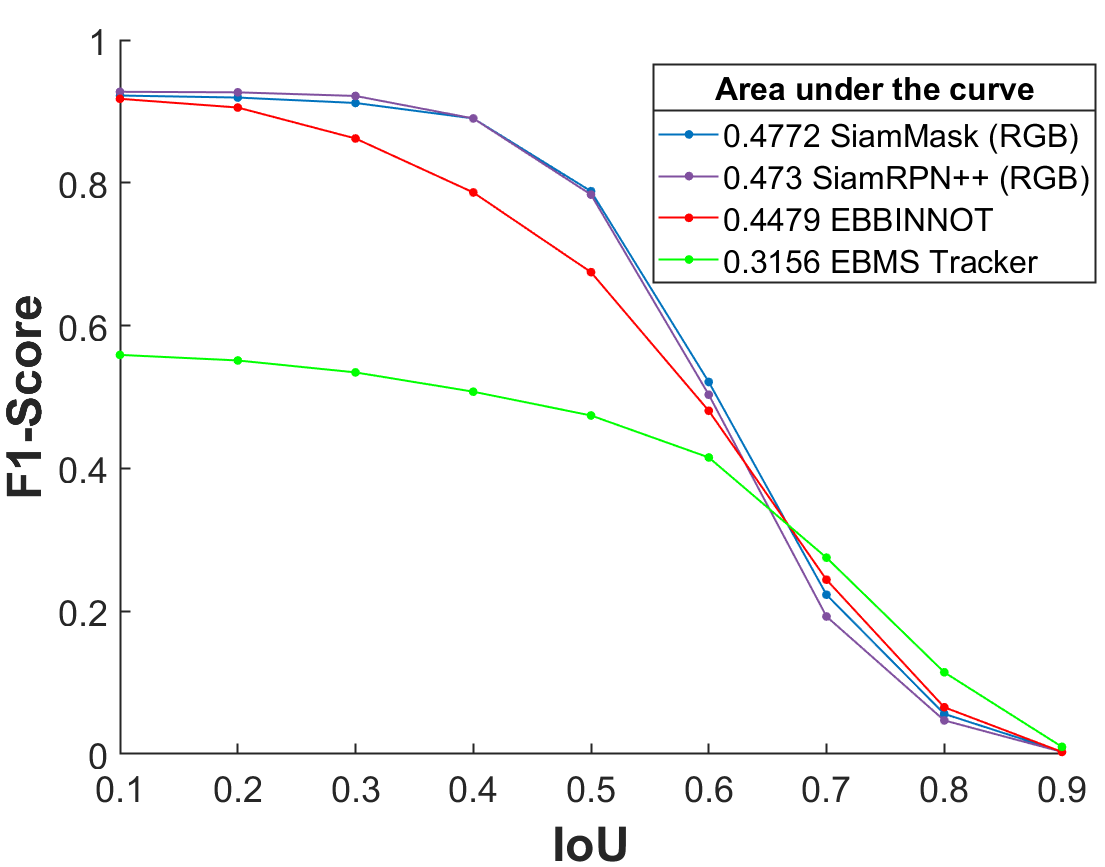} 
\caption{Comparison of the proposed EBBINNOT, SiamMask \cite{DBLP:Siammask}, SiamRPN++ \cite{SiamRPN++_Li_2019_CVPR} and Event-Based Mean-Shift (EBMS) \cite{delbruck2013robotic} at Site 3}
\label{fig:SiamMaskcomp}
\end{figure}
\subsection{Comparison to state-of-the-art } 
\label{sec:statecompare}
In this section, we report the performance of the proposed EBBINNOT compared to the frame-based state-of-the-art trackers, namely SiamMask \cite{DBLP:Siammask} and SiamRPN++ \cite{SiamRPN++_Li_2019_CVPR}, and event-based state-of-the-art approach EBMS. %We report the data collection and preparation process, which involves recording both event and RGB data simultaneously at site 3 shown in Figure \ref{fig:Locations}. 
Since $2/3^{rd}$ of data recorded at site 3 was used to train NNDC model, as stated in Table \ref{tab:dataset_statistics_overview}, we used remaining $1/3^{rd}$ of data for evaluation. Note that the original RGB dataset collected simultaneous with event-camera is used as input to SiamMask and SiamRPN++ and corresponding outputs are referred as SiamMask-RGB and SiamRPN++-RGB in this section.

%Since $2/3^{rd}$ of data recorded at site 3 was used to train NNDC model, as stated in Table \ref{tab:dataset_statistics_overview}, we used remaining $1/3^{rd}$ of data for evaluation. Note that original RGB dataset with FoV $90^\circ$ is used as input to SiamMask and corresponding output is referred as SiamMask-RGB in this section.
\par

\begin{figure}[t]
\centering
\subfloat[Occlusion scenario.] {\includegraphics[width=9cm]{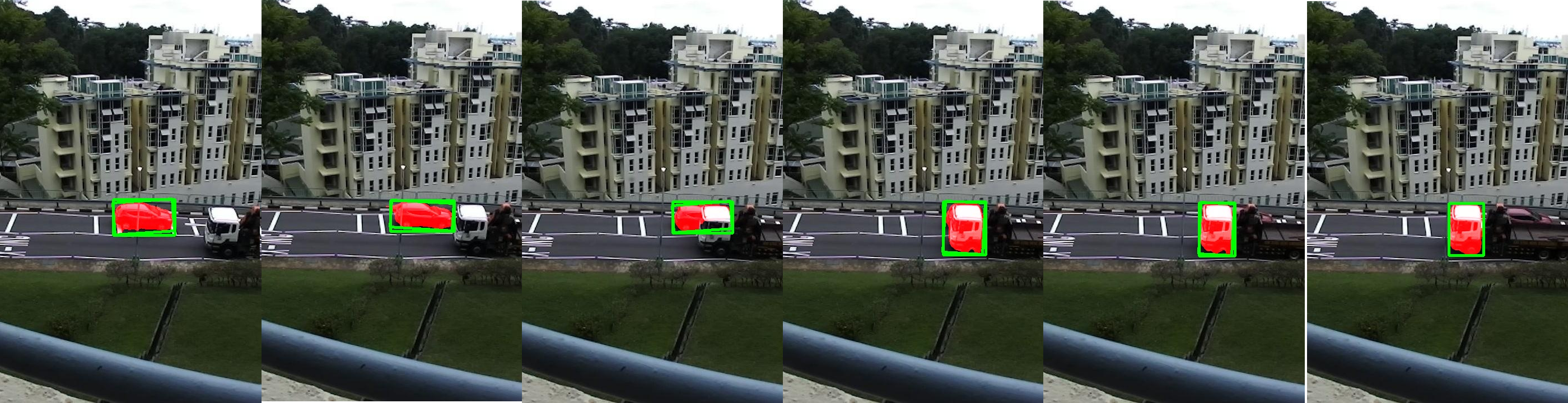}
\label{fig:SiamMask_Overlap_Sequence}}
\vspace{0.1em}
\subfloat[Background noise I.] {\includegraphics[width=9cm]{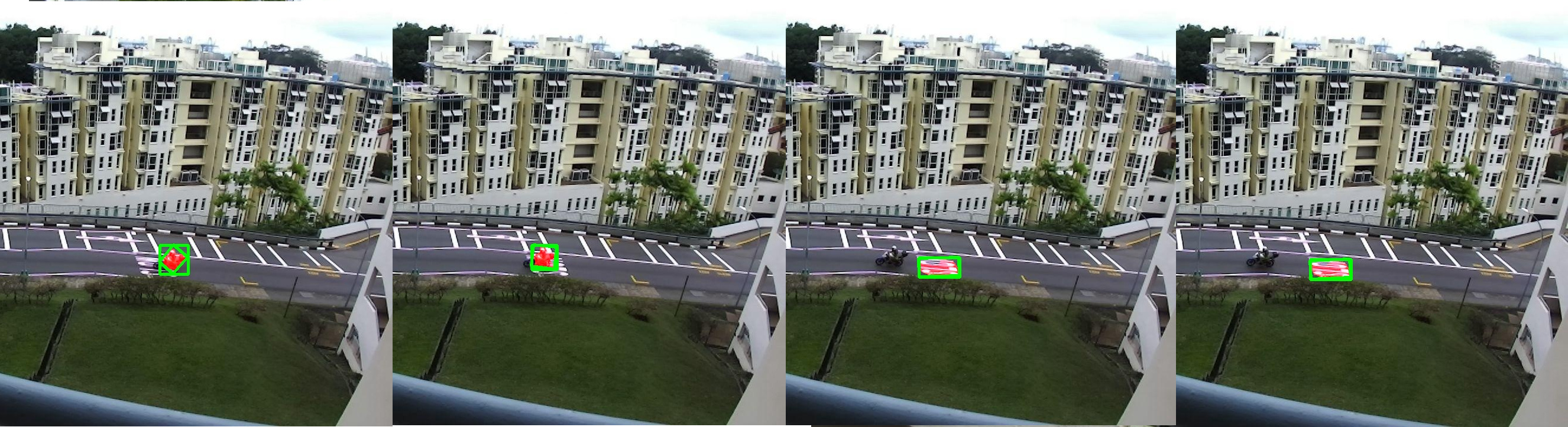}
\label{fig:SiamMask_Latching}}
\vspace{0.1em}
\subfloat[Background noise II.]
{\includegraphics[width=9cm]{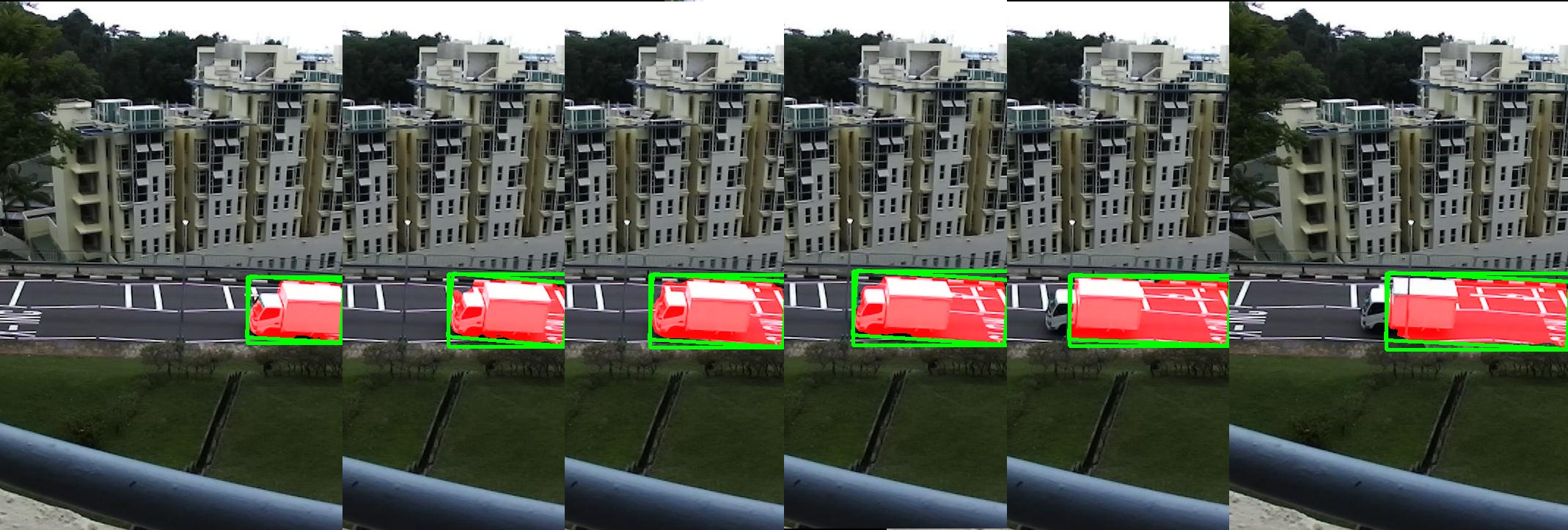}
\label{fig:SiamMask_Initialization_Scenario}}
\caption{Tracking performance of SiamMask under challenging scenarios. It fails to track either to background road markings or occluding objects becoming part of its online object learning representation}
\label{fig:SiamMask_failed_cases}
\end{figure}

Figure~\ref{fig:SiamMaskcomp} shows the F1 scores at various $\rm{IoU_{th}}$ for test recording. Even though Siamese DNNs perform marginally better than EBBINNOT due to their inherent use of similarity matching and our provision of ground truth initialization, EBBINNOT makes use of its own RPs for tracking and can make a good case for a practical system at lower IoUs. Mainly we observed a few scenarios when the background road markings and footpath patterns of the scene became part of the RGB object representation in Siamese networks as shown in Figure.~\ref{fig:SiamMask_failed_cases}, causing missed tracks.

Further investigation of performance of EBBINNOT and Siamese-DNNs on EBBI was studied and included in Sec. 4 of the \emph{Supplementary Material}\cite{ebbinnot-supplement}.

Overall, our proposed tracker comfortably outperforms the multi-object EBMS tracker while being on-par with Siamese DNN trackers. We attribute this to the need for re-scaling the frame size by half to $120\times90$ (previously noted in Section \ref{subsec:rpn_comparisons} as well), and consequently the NNDC model did not always pick a compact bounding box for some object categories. This drawback remains to be addressed in future works using techniques such as transfer learning.

\subsection{Computational Cost}
\label{subsec:Computational_Cost}

\begin{figure}[t]
\scriptsize
\centering     %%% not \center
\includegraphics[width=6cm]{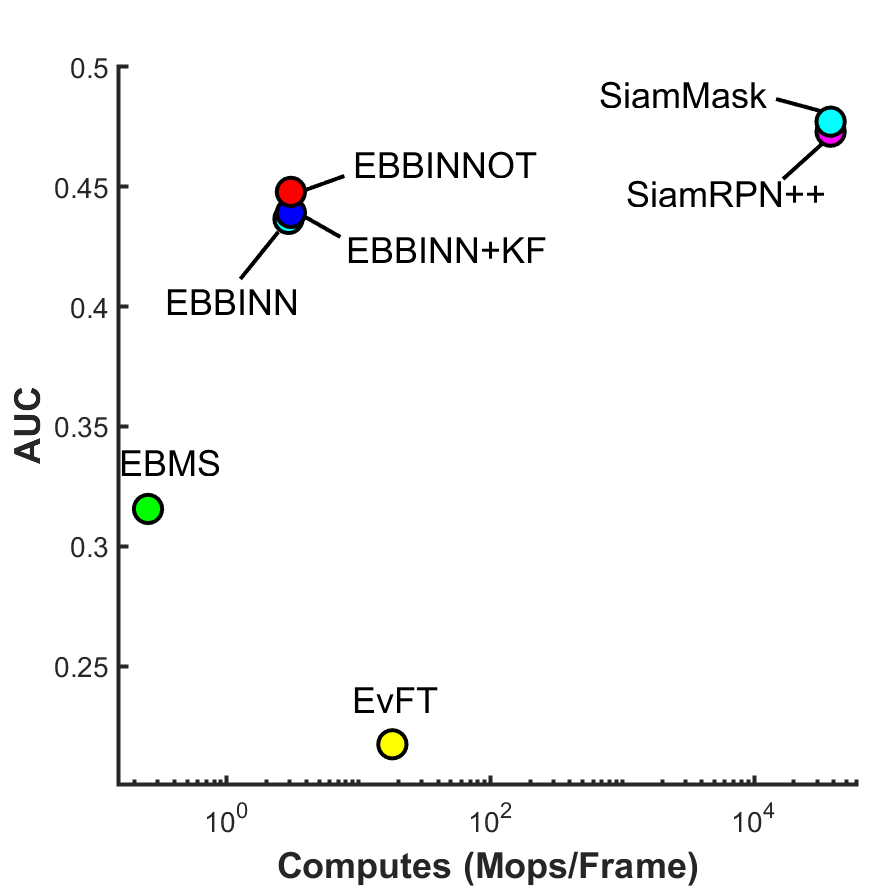}\\
(a)\\
\includegraphics[width=6cm]{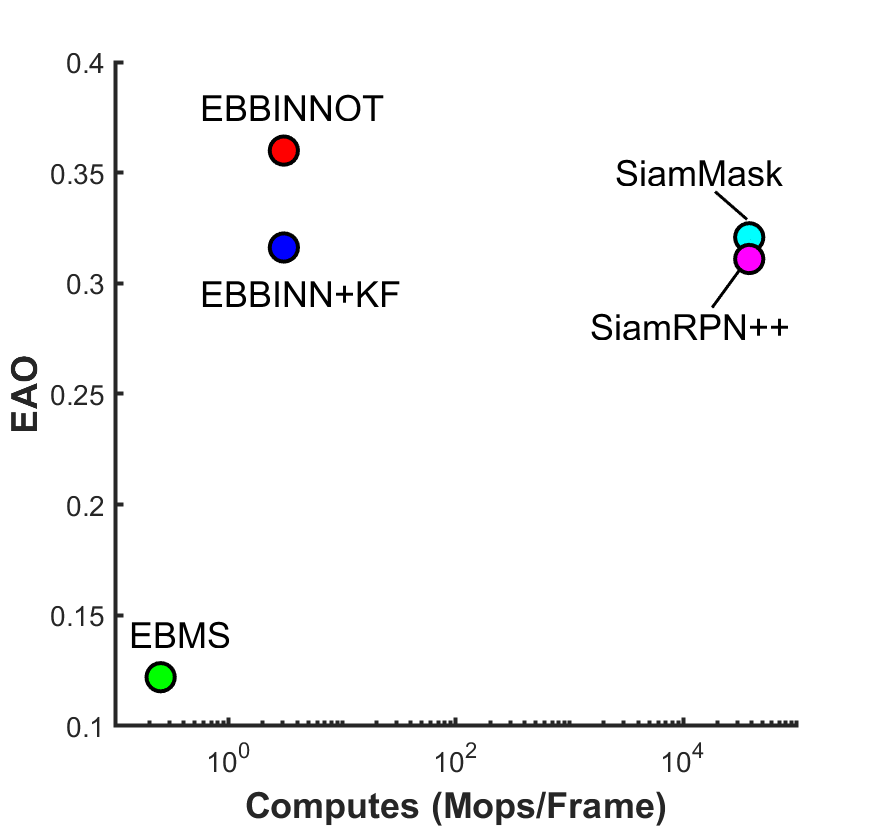}\\
\textcolor{red}{(b)}\\
 \caption{(a) Performance (measured as area under F1-curve - AUC) and computation cost comparison of the proposed EBBINNOT and other equivalent DVS trackers (b) Performance (measured by the expected average overlap (EAO) and computation cost comparison of the proposed EBBINNOT and other equivalent DVS trackers}
\label{fig:CC_AUC_comparison}
\end{figure}

% \begin{table}[]
% \centering
% \caption{Computational Count Estimation for Tracking Systems (Mops/Frame)}
% \begin{tabular}{ccccc}
% \hline
% EBMS & EvFT-KF & SiamMask & EBBINN-KF & EBBINNOT\\
% \hline\\
% XX & YY & ZZ & AA & BB\\
% \hline
% \end{tabular}
% \label{tab:trackerComputes}
% % \FloatBarrier
% \end{table}

% \begin{comment}
% \begin{table}[]
% \centering
% \caption{Computational Count Estimation: C{KF} vs C_{OT} }
% \begin{tabular}{ccccc}
% \hline
% \textbf{\begin{tabular}[c]{@{}c@{}}Recording\\ Site\end{tabular}} & \textbf{\begin{tabular}[c]{@{}c@{}}No. of\\ Recordings\end{tabular}} & \textbf{\begin{tabular}[c]{@{}c@{}}OT\\ Estimate\end{tabular}} & \textbf{\begin{tabular}[c]{@{}c@{}}KF\\ Estimate\end{tabular}} & \textbf{KF:OT} \\
% \hline
% Site 1             & 4                     & 119                   & 698                   & 6              \\
% Site 2             & 4                     & 351                   & 2472                  & 7              \\

% \hline
% \textbf{Average}            & \textbf{8 Recordings}                     & \textbf{235}                   & \textbf{1585}                  & \textbf{6.5}\\
% \hline
% \end{tabular}
% \label{tab:trackerComputes}
% % \FloatBarrier
% \end{table}
% \end{comment}
We compare the computational cost of six methods here: EBMS, EvFT, SiamMask, EBBINN-KF and the proposed EBBINNOT as shown in Figure \ref{fig:CC_AUC_comparison} The detailed equations for these calculations are shown in \emph{Supplementary Material}\cite{ebbinnot-supplement}.

For the trackers, the average number of computations per frame performed by KF-Tracker ($C_{KF}$) and OT ($C_{OT}$) were estimated following eq. S1 and eq. S3 in the \emph{Supplementary Material}\cite{ebbinnot-supplement} respectively, and these results were verified to be close to the actual computation count obtained by incrementing a counter with count weighted by computations in a step at run time, with an error margin of $\pm0.01\%$. Averaged across $8$ recordings at $2$ sites, KF-Tracker performs $\approx6.5\times$ more computations  as compared to the OT. However, in the whole EBBINNOT pipeline, this part is dwarfed by the NNDC block for RP as shown earlier in Fig. \ref{fig:comp_cost_pie_chart}.

\par
Based on the SiamMask architecture presented in \cite{DBLP:Siammask}, computations and memory usage are calculated layer-by-layer and then summed considering all network parameters and input dimensions. Total computations and memory requirements were deduced to be $\approx38000M$ operations per frame and $\approx 157MB$ respectively.

\par
The computations for EBBINN are obtained from Eq. \ref{eq:comp_nndc} where the values of $\alpha_T=0.57$ and $\overline{n_{\rm{RP}}}=2.38$ are estimated from the traffic dataset. It can be seen that SiamMask uses $\approx12427
\times$ more computes per frame and demands $\approx1450\times$ more memory than EBBINNOT, due to its Siamese-based deep neural network architecture. Since, SiamRPN++ differs from SiamMask only in the last layer of its architecture, we approximate it to have similar computational and memory requirements as SiamMask. Therefore, EBBINNOT offers a fair advantage in terms of total computation and memory usage.
\par

% The computes per frame for EBMS can be calculated using eq.~\ref{eq:EBMS}, as calculated in \cite{AuthorsEBBIOT}. 
% \begin{align}
% \label{eq:EBMS}
%     C_{EBMS} =& \overline{N}\times [9~\overline{CL}^2+{(169+16~\gamma_{merge})}~\overline{CL} +11]\notag\\
%     M_{EBMS} =& 408 CL_{max}+ 56
% \end{align}
Assuming the past $10$ positions of cluster for the current velocity calculation, $CL_{max}=8$ and for our dataset, $\overline{CL}\approx 2$, $\gamma_{merge}\approx0.1$ and $\overline{N_F}\approx 650$, EBMS requires 252 kops per frame as estimated in \cite{AuthorsEBBIOT} which is 12$\times$ lower than EBBINNOT, and a memory of $3.32$KB, which is nearly negligible. The proposed EBBINNOT, however,  significantly outperforms EBMS as shown in Figures~\ref{fig:filtering_RP_track_Comparison}(c) \& \ref{fig:SiamMaskcomp}. This performance gain comes at the cost of slightly higher computations and memory usage. Overall, out of the three approaches considered here, EBBINNOT offers the best trade off between performance and computational complexity, even though Siamese DNNs perform marginally better.

Lastly, coming to the case of EvFT RP, most computations here (over $90\%$) are consumed by two EM flows of the feature tracker. The RP creation process adds negligible overhead to the tracking process. We estimated the number of computations required to implement the tracker + RP to be of the order of $1$ Mops per call of the EM functions by theoretical calculation and experimental verification by profiling the MATLAB code provided in \cite{Kostas-flow}. For a given recording in our dataset comprising $\approx 17M$ events spanning $\approx16$ minutes,  the average number of computes per frame for  EvFT is $\approx18$ Mops, which is $\approx 5.89 \times$ more than that required by EBBINN.

Apart from the AUC obtained from F-1 curves, we also compare the tracking results using a metric of expected average overlap (EAO) commonly used in object tracking\cite{vot2015}. In this method, the overlap between ground truth and its longest matching track are evaluated for every frame with the overlap set at zero for frames where the track is lost. The expected value is obtained by averaging across tracks of different length from the entire dataset. The results plotted in Fig. \ref{fig:CC_AUC_comparison}(b) shows a similar trend where the DL trackers and EBBINNOT outperform EBMS. EvFT is not included in this comparison since it does not generate tracks but is used only as a region proposal. Interestingly, while the KF tracker operated on region proposals from EBBINN are worse than the DL trackers, EBBINNOT outperforms them due to the heuristics to handle occlusion.
\section{Discussion}
\label{sec:discussion}
%\input{05_01_tracks_figtex}

% \begin{figure}[t]
% \centering     %%% not \center
% \includegraphics[width=9cm]{img/plots/cclToNNDCImprovement.png}
%  \caption{Weighted F1 scores for tracker (OT or KF) run on CCL Vs. CCL + NNDC RP}
% \label{fig:cclToNNDCImprovement}
% \end{figure}

%\begin{figure}[]
%\scriptsize
%    \centering
%  \begin{tabular}{@{}c@{}}
%    \includegraphics[width=9cm]{img/plots/cclToNNDCImprovement.png} \\[\abovecaptionskip]
%  \end{tabular}
%\begin{tabular}{|c|c|c|c|}
%                \hline
%                Curve Name & EBBI+CCL+OT & EBBI+CCL+KF & EBBI+CCL+NNDC+OT \\ \hline
%                AUC & 0.2396 & 0.2088 & 0.3938 \\ \hline
%\end{tabular}
%    \caption{Weighted F1 scores for tracker (OT or KF) run on CCL Vs. CCL + NNDC RP}
%    \label{fig:cclToNNDCImprovement}
%\end{figure}
%\FloatBarrier
\subsection{Hardware Implications}
We have earlier described the computational complexity of the proposed EBBINNOT and compared with the other approaches such as Siam Mask or EvFT. To get a more concrete idea about the energy requirement, we consider two separate implementations using a low-power microprocessor (uP)\cite{Ambiq} or a custom chip (ASIC). The lowest energy per operation for the uP is estimated to be $\approx 6$ pJ\cite{Ambiq}. Hence, we can estimate the energy requirement for SIAMMASK, EBBINNOT and EvFT to be $\approx 228$ mJ, $18.3$ $\mu$J and $107.7$ $\mu$J per frame respectively. Since the major computational burden of both SIAMMASK and EBBINNOT are attributed to neural networks, we can use the recently popular concept of in-memory computing (IMC)\cite{IMC-1,IMC-2,IMC-3} based chip to gain further energy efficiencies. Considering an average energy per operation of $475$ fJ based on recent works\cite{IMC-1,IMC-2,IMC-3}, we can estimate the energy per frame for SIAMMASK and EBBINNOT to be $18.1$ mJ and $1.44$ $\mu J$ respectively. Note that here we have ignored the fact that parts of the computation (e.g. tracker) may not be optimized by the IMC architecture since these computations are a small fraction of the total computes as shown in Fig. \ref{fig:comp_cost_pie_chart}.
\subsection{Repeatability of results for recordings with other NVS - CeleX}
The proposed flow consisting EBBI creation, median filtering, NNDC and OT was also verified for repeatability on recordings from the CeleX~\cite{guo2017live} camera. We collected a total of $35$ recordings at different times from a single location and divided them in the ratio of 5:2 for training and testing. After reviewing the size distribution of the objects from different classes obtained from the GT annotations and comparing it with the distribution from earlier recordings, we settled to resize the images by a factor of $3.33$ to $384\times240$ from $1280\times800$. 
\par
Due to the camera's invalid polarity output at some points, only 1B1C images were stored for training the NNDC model. Thus, during training and testing, the RP input to NNDC had a size of $42\times42\times1$ and the output BB coordinates were also scaled according to the new image size. We balanced the training data by augmentation and further removed the excess number of examples in classes like cars because they often appeared in the field of view. 
\begin{table}[!]
\caption{Classification Scores for testing videos recorded on CeLeX}
\centering
\begin{tabular}{lcc}
\hline
{\color[HTML]{000000} \textbf{Category}} & {\color[HTML]{000000} \textbf{per sample (\%)}} & {\color[HTML]{000000} \textbf{per track (\%)}} \\ \hline
{\color[HTML]{000000} Car/Van} & {\color[HTML]{000000} 93.38} & {\color[HTML]{000000} 95.61} \\
{\color[HTML]{000000} Bus} & {\color[HTML]{000000} 96.4} & {\color[HTML]{000000} 95.65} \\
{\color[HTML]{000000} Bike} & {\color[HTML]{000000} 89.38} & {\color[HTML]{000000} 96.3} \\
{\color[HTML]{000000} Truck} & {\color[HTML]{000000} 50.57} & {\color[HTML]{000000} 80} \\ \hline
{\color[HTML]{000000} \textbf{Unbalanced accuracy}} & {\color[HTML]{000000} \textbf{91.64}} & {\color[HTML]{000000} \textbf{95.27}} \\
{\color[HTML]{000000} \textbf{Balanced accuracy}} & {\color[HTML]{000000} \textbf{82.43}} & {\color[HTML]{000000} \textbf{91.89}} \\ \hline
\end{tabular}

\label{tab:CeLeX scores}
\end{table}
\begin{figure}[]
\centering     %%% not \center
\includegraphics[width=6cm]{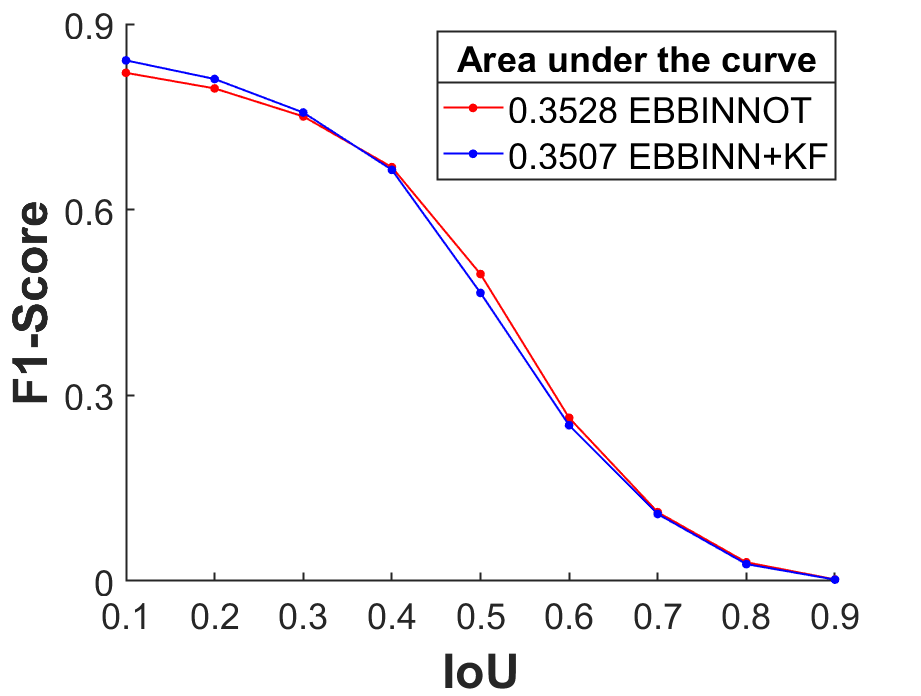}
 \caption{Weighted F1 scores for CeleX recordings showing similar performance to the DAVIS recordings}
\label{fig:CelexF1}
\end{figure}

%\begin{figure}[]
%\scriptsize
%    \centering
%  \begin{tabular}{@{}c@{}}
%    \includegraphics[width=9cm]{img/plots/celexTrackerPerfComparison.png} \\[\abovecaptionskip]
%  \end{tabular}
%\begin{tabular}{|c|c|c|}
%                \hline
%                Curve Name & EBBI+CCL+NNDC+KF & EBBI+CCL+NNDC+OT \\ \hline
%                AUC & 0.3507 & 0.3528 \\ \hline
%\end{tabular}
%    \caption{Weighted F1 scores for CeleX recordings}
%    \label{fig:CelexF1}
%\end{figure}
%\FloatBarrier
The model was trained on the same configurations explained in Section \ref{subsec:rpn_comparisons} and the best model yielded the classification results shown in Table~\ref{tab:CeLeX scores}. Moreover, the detection performance was checked after running the actual 10 testing videos on the entire setup and the generated annotations were rescaled back to the original sensor dimensions by multiplication with 3.33. We calculated the weighted F1 scores shown in Figure~\ref{fig:CelexF1}, and as expected, the performance for these recordings are comparable to the performance results of the proposed flow for DAVIS240C recordings, thus proving repeatability and reproducibility of the results for different recordings from different neuromorphic vision sensors.

\section{Conclusion}
\label{sec:conclusion}
This paper proposes a new hybrid event-frame pipeline called EBBINNOT for the IoT based traffic monitoring system using a stationary DVS. EBBINNOT creates an event-based binary image and uses median filtering to remove noise. The combination of EBBI and median filtering acts like a generalized version of nearest neighbour filtering but results in much reduced hardware cost. The output is then sent to a connected component labelling based region proposal network followed by NNDC for merging fragmented proposals, predicting their correct sizes and class categories. The modified proposals are then passed to an overlap based tracker having tracking/locked state trackers, heuristics for handling occlusion and other simplified methods inspired from Kalman Filter. All the mentioned blocks in EBBINNOT are completely optimized for computational costs. EBBINNOT requires $\approx3.057M$ operations per frame, almost $12427\times$ less than the state-of-the-art purely frame-based Siamese DNN trackers while providing on-par performance  
%and even outperforms it in tracking performance by AUC of $\approx0.12$ 
calculated on the simultaneously collected events and RGB data. Further, this system also shows a substantial improvement over the purely events-based approach called EBMS\cite{delbruck2013robotic} with tracking performance difference of AUC $\approx0.14$, though requiring $ 12\times$ more computations. Compared to more recent event-based feature trackers\cite{Kostas-flow}, the proposed method requires $\approx 5.89\times$ lesser computes while delivering similar tracking performance.

EBBINNOT also achieves an overall balanced track accuracy of $92.70\%$ on recordings from three sites spanning more than five hours. Our results show the great promise offered by neuromorphic vision sensors in monitoring applications required by IoT. Future work will focus on hardware realizations of the described pipeline and extending this work to monitoring human activity in crowded areas. New concepts from object tracking\cite{Karunasekera2019,Luiten2020} in computer vision can also be integrated in this framework. Further, in the context of traffic monitoring, this work will be extended to cover other environmental conditions such as rain or fog where the high dynamic range of DVS may provide an advantage.

\Urlmuskip=0mu plus 1mu\relax
%\raggedright
\bibliographystyle{IEEEtran}

\bibliography{JETCAS2018}
\end{document}